\documentclass[times,twocolumn,final]{elsarticle}
\usepackage{medima_modified}

\usepackage{framed,multirow}

\usepackage{amssymb}
\usepackage{latexsym}

\usepackage{url}
\usepackage{xcolor}

\usepackage{hyperref}
\usepackage{textcomp}

\definecolor{newcolor}{rgb}{.8,.349,.1}

\usepackage[cmex10]{amsmath}

\usepackage{bm}
\journal{Medical Image Analysis}

\begin{document}

\verso{Qiao Zheng \textit{et~al.}}

\begin{frontmatter}

\title{Explainable cardiac pathology classification on cine MRI with motion characterization by semi-supervised learning of apparent flow}%

\author[1]{Qiao \snm{Zheng}\corref{cor1}}
\cortext[cor1]{Corresponding author: 
  Tel.: +33-492385024;  
  fax: +33-492387669;
  Inria Epione, 2004 route des Lucioles BP 93, 06902 Sophia Antipolis Cedex, France}
\ead{qiao.zheng@inria.fr}

\author[1]{Herv\'{e} \snm{Delingette}}

\author[1]{Nicholas \snm{Ayache}}

\address[1]{Universit\'{e} C\^{o}te d'Azur, Inria, France}

%\received{1 November 2018}
%\finalform{10 November 2018}
%\accepted{13 November 2018}
%\availableonline{15 November 2018}
%\communicated{S. Sarkar}

\begin{abstract}
%%%
We propose a method to classify cardiac pathology based on a novel approach to extract image derived features to characterize the shape and motion of the heart. An original semi-supervised learning procedure, which makes efficient use of a large amount of non-segmented images and a small amount of images segmented manually by experts, is developed to generate pixel-wise apparent flow between two time points of a 2D+t cine MRI image sequence. Combining the apparent flow maps and cardiac segmentation masks, we obtain a local apparent flow corresponding to the 2D motion of myocardium and ventricular cavities. This leads to the generation of time series of the radius and thickness of myocardial segments to represent cardiac motion. These time series of motion features are reliable and explainable characteristics of pathological cardiac motion. Furthermore, they are combined with shape-related features to classify cardiac pathologies. Using only nine feature values as input, we propose an explainable, simple and flexible model for pathology classification. On ACDC training set and testing set, the model achieves 95\% and 94\% respectively as classification accuracy. Its performance is hence comparable to that of the state-of-the-art. Comparison with various other models is performed to outline some advantages of our model.
%%%%
\end{abstract}

\begin{keyword}

%% Keywords
\KWD Cardiac pathology\sep Classification\sep Cine MRI\sep Motion\sep Deep learning\sep Semi-supervised learning\sep Neural network\sep Apparent flow
\end{keyword}

\end{frontmatter}

%\linenumbers

\section{Introduction}
\label{sec1}

Cine magnetic resonance imaging (cine MRI) is widely used in the clinic as an approach to identify cardiac pathology. For both the patients and the clinicians, there is hence a great need for automated accurate cardiac pathology identification and classification based on MRI images as mentioned in \cite{Rueckert:2016} and \cite{Comaniciu:2016}, as well as in the myocardial infarct classification challenge run at the STACOM workshop in 2015 (\cite{Suinesiaputra:2018}). Recently, the state-of-the-art cardiac pathology classification methods extract various features from MRI images and perform classification based on these features. Despite the great results achieved so far, there are still some aspects that need to be further explored.

First, most classification models, including the state-of-the-art models, take many feature values together as input to a single or a group of machine learning classifiers (e.g. \cite{Khened:2017}, \cite{Khened:2018}, \cite{Wolterink:2017}, \cite{Cetin:2017}, \cite{Isensee:2017}), and output the predicted probability distribution over several classes. Like many other machine learning methods, or more specifically like most deep learning methods, these classification models are not easy to interpret. On the one hand, most of the models contain at least hundreds of parameters and it is impractical to examine and explain the role of each parameter. On the other hand, as many features are used simultaneously, it is hard to tell in a straightforward manner which feature value contributes to the identification of which category. This drawback on explainability causes many problems as pointed out in \cite{Holzinger:2017}. For instance, the lack of explainability is a significant hurdle for their widespread adoption in the clinic despite their performance. Moreover, under the new European General Data Protection Regulation, it may also generate legal issues in business, as companies are required to be able to explain why decisions have been made by their models upon demand. Hence we propose a simple classification model with 9 input features and 14 parameters in total such that the role and contribution of each feature or parameter are clear and explainable.

Second, in terms of data availability in medical image analysis, we usually have access to a large amount of unlabeled data and a small amount of labeled data. How to make good use of the available data to train automatic methods remains an open question (\cite{Weese:2016}). Semi-supervised learning appears to be a powerful approach to tackle this challenge in general (\cite{Bai:2017:2}, \cite{Gu:2017}, \cite{Cheplygina:2018}). In this paper, while cardiac motion is estimated in a flow-based manner like in many other methods (\cite{Gao:2016}, \cite{Parajuli:2017}), we extend it as a semi-supervised learning method to train a network for apparent flow generation, using the dataset of Automatic Cardiac Diagnosis Challenge (ACDC) of MICCAI 2017 (\cite{Bernard:2018}), for which the ground-truth segmentation mask is only available for 2 time frames. Although the percentage of the segmented frames in the dataset is small, making efficient use of their segmentation masks in training is essential for the generated flow to have better consistency. In particular, with the supervision of the masks in training, we show that cardiac structures are better preserved after warping by the generated flow.

Third, the state-of-the-art classification methods most exclusively focus on features extracted at two instants: the instants of end-diastole (ED) and end-systole (ES). The other instants are often ignored in pathology classification. For example, in the ACDC challenge, 3 out of the 4 cardiac pathology classification methods, including \cite{Khened:2017} (as well as its updated version \cite{Khened:2018}), \cite{Wolterink:2017} and \cite{Cetin:2017}, use only features based on ED and ES. The authors of \cite{Isensee:2017} propose the only method in the ACDC challenge which explores the instants other than ED and ES by quantifying the volume change and by measuring the LV-RV dissynchrony. Yet much information about cardiac motion (e.g. how individual myocardial segments move) is still excluded from the extracted features. While more and more research efforts are put on cardiac motion estimation (e.g. \cite{Qin:2018}, \cite{Qin:2018:2}, \cite{Xue:2018}, \cite{Yang:2017}, \cite{Yan:2018}) and cardiac disease assessment via motion analysis (e.g. \cite{Gilbert:2017}, \cite{Dawes:2017}, \cite{Lu:2018}), we propose to explore the impact of specific motion features to learn the detection of cardiac pathologies by extracting some useful time series of simple and straightforward features from cine MRI image sequences. Ideally, the resulting time series should be both informative enough to be used for classification and intuitive to be understood by a physician.

In this paper, we propose a novel and explainable method to classify a subset of cardiac pathologies using deep learning of cardiac motion (in the form of apparent flows) and shape. Our main contribution is threefold:\\
\textbullet \ \textbf{Semi-supervised learning of flow}: a novel semi-supervised learning method is applied to train a neural network model, which outputs apparent flows given two MRI images from the same 2D+t cine MRI image sequence. This allows to learn the motion as apparent flows efficiently from both segmented and non-segmented image data. \\
\textbullet \ \textbf{Motion-characteristic features}: combining the apparent flows across time with cardiac segmentation, time series of the radius and thickness of myocardial segments are extracted to describe cardiac motion. As features, they are easy to interpret and allow to characterize different shapes and motions of cardiac pathologies.\\
\textbullet \ \textbf{Explainable classification model}: we train a set of 4 simple classifiers to perform binary classifications. Each classifier performs a logistic regression and takes no more than 3 feature values as input, which makes it very simple and easy to interpret. On the ACDC challenge training set and testing set, our model achieves 95\% and 94\% as classification accuracy respectively, which is comparable to the state-of-the-art.

\begin{figure*}[]
\centering
\includegraphics[width=15.5cm, height=8cm]{feature_method_overview2.pdf}
\caption{Overview of the feature extraction method: 1. Apparent flow generation given the ED frame and another frame on the same slice; 2. Cardiac segmentation on the ED and ES frames and division of the ED myocardium mask to 6 segments; 3. Extraction of the shape-related features, including the calculation of the volumes, volume ratios and myocardial thickness of a heart given the segmentation masks; 4. Extraction of motion-characteristic features, including the creation of segment radius and thickness time series given a slice with the corresponding apparent flow maps and segmentation mask.}
\end{figure*}

\section{Data}

\subsection{Dataset}
The proposed method is trained and evaluated on the ACDC challenge dataset, which consists of a training set of 100 cases and a testing set of 50 cases. The cine MRIs were acquired with a conventional SSFP sequence (\cite{Bernard:2018}). Most of the cases contain about 10 slices of short-axis MRIs. And the number of frames in the cases varies between 12 and 35. ACDC training set and testing set are respectively divided into 5 pathological groups of equal size (we cite below the properties of each group as provided on the website, though they are only roughly exact according to our measure and observation): \\
\textbullet \ dilated cardiomyopathy (DCM): left ventricle cavity (LVC) volume at ED larger than 100 $\mathit{mL/m^2}$ and LVC ejection fraction lower than 40\% \\ 
\textbullet \ hypertrophic cardiomyopathy (HCM): left ventricle (LV) cardiac mass higher than 110 $\mathit{g/m^2}$, several myocardial segments with a thickness higher than 15 mm at ED and a normal ejection fraction \\ 
\textbullet \ myocardial infarction (MINF): LVC ejection fraction lower than 40\% and several myocardial segments with abnormal contraction\\
\textbullet \ RV abnormality (RVA): right ventricle cavity (RVC) volume higher than 110 $\mathit{mL/m^2}$ or RVC ejection fraction lower than 40\% \\
\textbullet \ normal subjects (NOR) \\
Please note that the abnormal contraction mentioned in the characteristics of MINF is quite vague as a property. In addition, both MINF and DCM cases have low LVC ejection fractions. And sometimes, a myocardial infarction causes a dilated LVC (for which we should classify the case to MINF instead of DCM according to ACDC challenge). As we will present later, it is indeed a challenge to distinguish them.

For the cases of ACDC training set, expert manual segmentation for LVC, RVC and the left ventricular myocardium (LVM) is provided as ground-truth for all slices at ED and ES phases; all other structures in the image are considered as background. For the cases of ACDC testing set, no ground-truth information about classification or segmentation is available. For performance evaluation on the testing set, the predicted results of a model need to be submitted online.

\subsection{Notation}
In this paper, slices in image stacks are indexed in spatial order from the basal part to the apical part of the heart.  Given an image stack $S$, we denote $N_S$ the number of its slices. Given two values $a$ and $b$ between $0$ and $N_S-1$, we note $S[a,b]$ the sub-stack consisting of slices of indexes in the interval $[\mathit{round}(a), \mathit{round}(b)[$ ($round(a)$ is included while $\mathit{round}(b)$ is excluded).
%with $\mathit{round}$ the function rounding to nearest integer. For instance, if $S$ is a stack of $N_S$=$10$ slices of indices from 0 to 9, then $S[0.2N_S, 0.6N_S]$ is the stack consisting of slices of indices 2, 3, 4 and 5.

\begin{figure}[t]
\centering
\includegraphics[width=4.8cm, height=8cm]{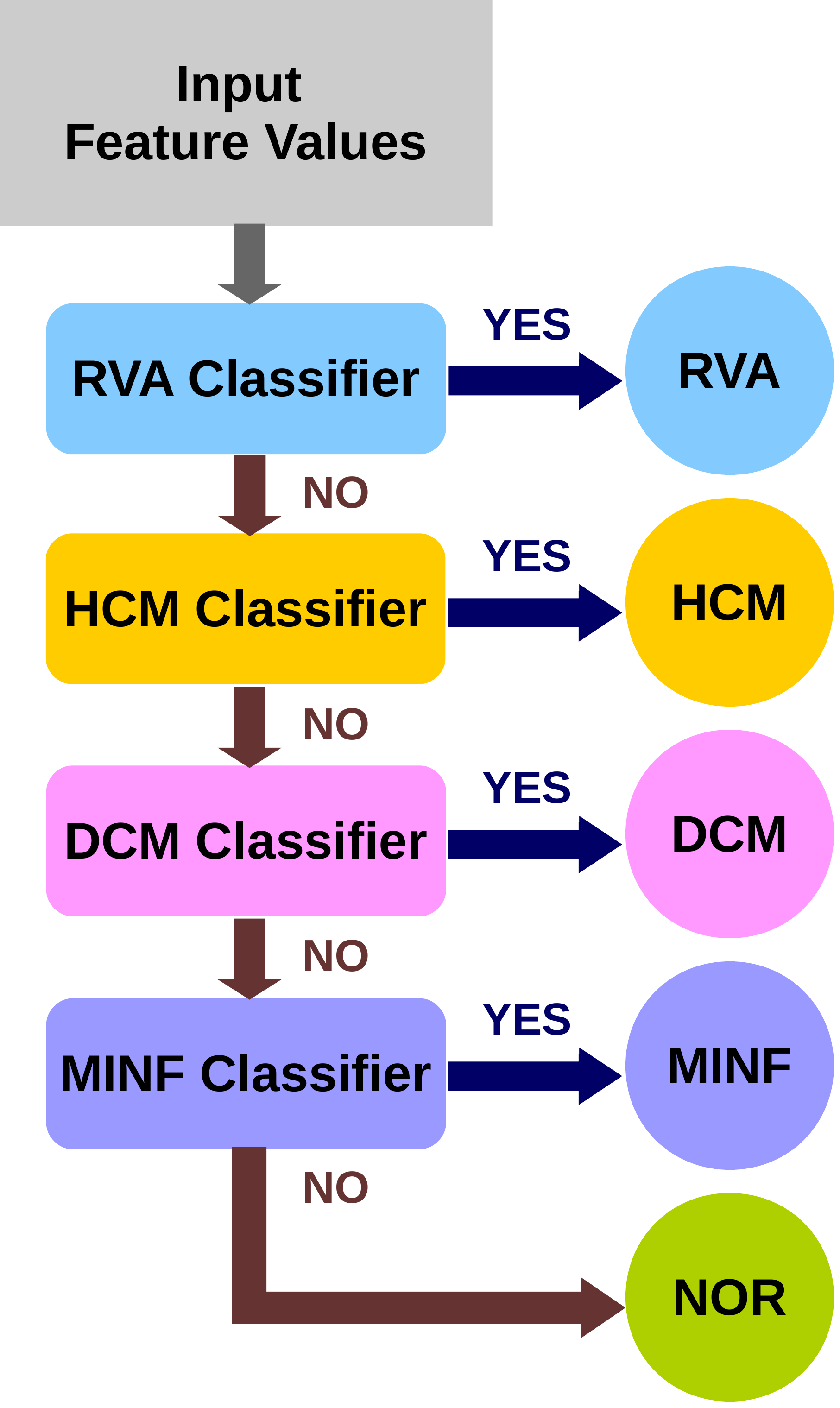}
\caption{Overview of the classification method: the 4 binary classifiers are applied in sequence to classify a case to RVA, HCM, DCM, MINF or NOR.}
\end{figure}

\section{Methods}

Our method mainly consists of two parts: feature extraction (Fig.1) and classification based on features (Fig.2). But the region of interest (ROI) needs to be determined first.

\subsection{Preprocessing: Region of Interest (ROI) Determination}

As a preprocessing step, the ROI needs to be determined on the original MRI images. Short-axis MRI images usually cover a zone much larger than that of the heart. To save memory usage and to increase the speed of apparent flow and segmentation methods, it is better to work on an appropriate ROI instead. For this purpose, we directly apply an existing ROI method: we use the trained ROI-net exactly as described in \cite{Zheng:2018} to define an ROI. Briefly speaking, the ROI-net is a variant of U-net (\cite{Ronneberger:2015}) for heart/background binary segmentation. It is applied on several middle slices on the ED image stack. As shown in \cite{Zheng:2018}, this ROI determination method is very robust and succeeds in all cases of the ACDC dataset. In the remainder of this paper, we only refer to the automatically cropped ROI of the images.

\subsection{Feature Extraction Step 1: Apparent Flow Generation}

\begin{figure}[]
\centering
\includegraphics[width=8.5cm, height=8cm]{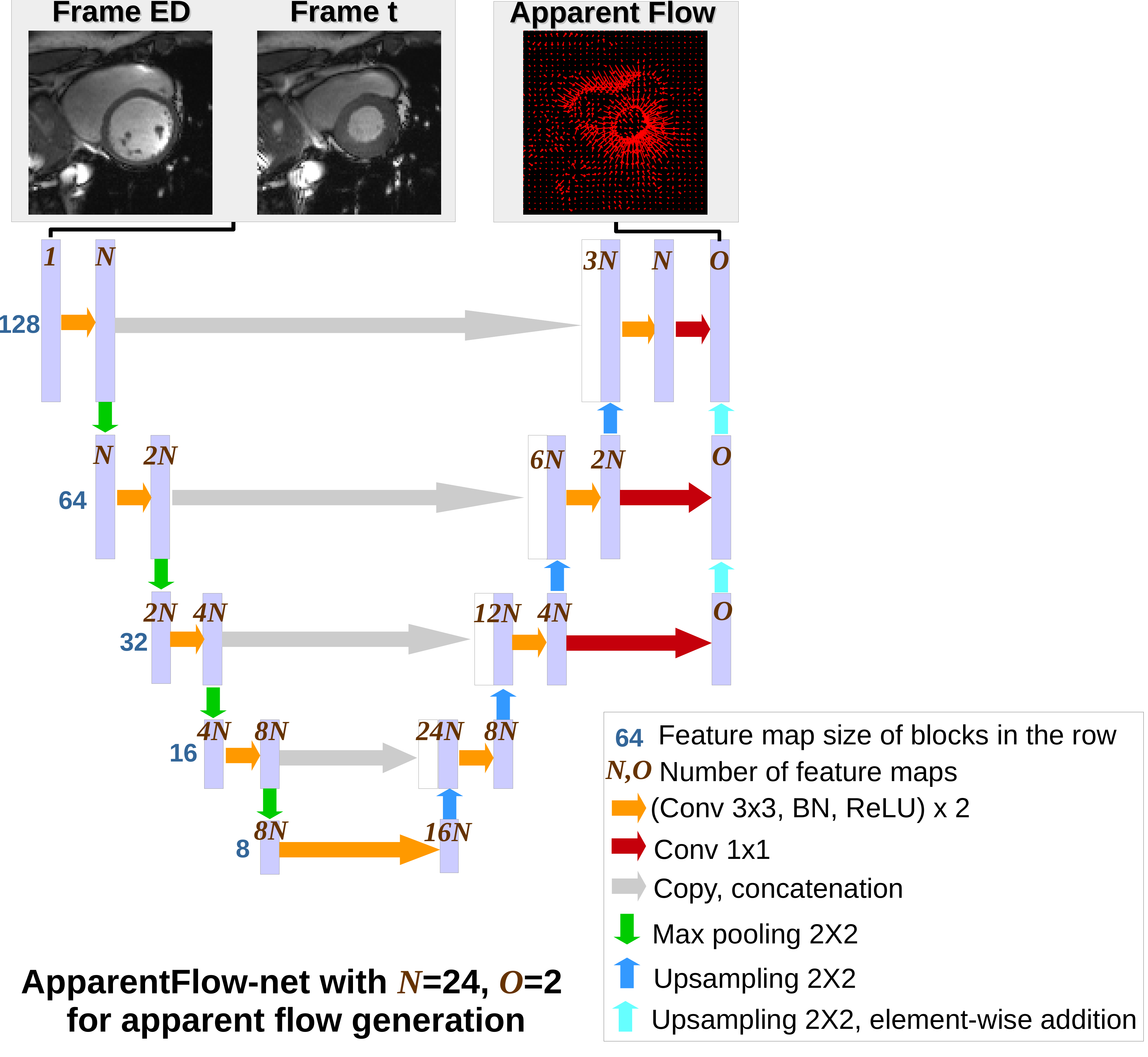}
\caption{ApparentFlow-net: for apparent flow generation. The output is a map of pixel-wise flow $\bm{F_{t}}$.}
\end{figure}

As shown in Fig.1, there are four steps for feature extraction. In this first step, the ApparentFlow-net, which is a variant of U-net (\cite{Ronneberger:2015}) as shown in Fig.3, is proposed. U-net, with the encoder-decoder structure consisting of layers of various sizes of receptive fields, can effectively integrate local and global information, which is necessary for the analysis of the shape and motion of the heart on MRIs. Previously, we successfully used some variants of U-net for cardiac segmentation (\cite{Zheng:2018}). So we expect a similar structure would also work for the estimation of cardiac motion. The ApparentFlow-net is applied to generate pixel-wise apparent flow given a pair of image frames on the same slice as input: the ED frame and another frame of index $t$ on the same slice. In other words, the generated apparent flow map is a displacement field of the slice between ED and instant $t$. In a later step, combined with the segmentation mask, we will extract cardiac motion features from the sequences of apparent flow maps on a slice. The details of this extraction are available in the sub-section 3.5. While there exists some researches that explore image registration (or equivalently, apparent flow) using unsupervised learning (e.g. \cite{Balakrishnan:2018}, \cite{Krebs:2018}, \cite{deVos:2017}, \cite{Li:2017}), we propose a semi-supervised learning approach to make efficient use of a large amount of non-segmented images and a small amount of images segmented manually by experts.

In general, the idea of representing motion by apparent flow is based on two assumptions. First, we assume that the pixel intensities of an object do not change much between the two frames. Second, it is assumed that neighboring pixels have similar motion. By observation, we find that these assumptions usually hold on the slices located below the base and above the apex with some margin. This is due to the limited out-of-plane motion on these slices (this is less the case for the slices around the base and the apex). Hence ApparentFlow-net is trained and applied on the middle slices only.

If we note $I_{\mathit{ED}}(\bm{P})$ and $I_t(\bm{P})$ the pixel intensity of the two input frames of ApparentFlow-net at position $\bm{P}=(x, y)$, according to the first assumption above, ApparentFlow-net should generate an apparent flow map $\bm{F_{t}}$ with $\bm{F_{t}}(\bm{P}) = (F_{t}^x(\bm{P}), F_{t}^y(\bm{P}))$ between ED and $t$ enabling image reconstruction such that the following intensity discrepancy is minimized:
\begin{equation}
L_{\mathit{IMG}}(\bm{F_t}) = \sum_{\bm{P}}\bigg(I_{\mathit{ED}}(\bm{P}) - I_{t}\big( \bm{P}+\bm{F_t}(\bm{P}) \big)\bigg)^2
\end{equation}

Meanwhile, the flow should also preserve the regularity of the motion of neighboring pixels according to the second assumption above. While there are already some methods in the community to impose diffeomorphisms (e.g. demon's algorithm as in \cite{Pennec:1999}, LDDMM as in \cite{Hernandez:2008}), we propose a simple one to only discourage the occurrence of the extreme situations such as the crossing between two adjacent pixels or rotations greater than $90^\circ$(Fig.4). As long as these unrealistic motion patterns do not appear, there is no penalty on the regularity at all and the network is free to generate whatever flow without being influenced by the regularity constraint. More precisely, let us note $W_{\bm{F_{\mathit{t}}}}$ as the warping function such that $W_{\bm{F_{\mathit{t}}}}(\bm{P}) = \bm{P} + \bm{F_{\mathit{t}}}(\bm{P})$. For two adjacent pixels $\bm{P} = (x, y)$ and $\bm{P^{x+}} = (x+1, y)$ in a row, we want the warped pixel $W_{\bm{F_{\mathit{t}}}}(\bm{P^{x+}})$ to stay on the right of the warped pixel $W_{\bm{F_{\mathit{t}}}}(\bm{P})$ (similarly for the adjacent pixels $\bm{P}$ and $\bm{P^{y+}} = (x, y+1)$ in a column) (see Fig.4). Otherwise, we say that a crossing on the x-components (y-components) of the flow pairs occurs and a penalty should apply. This translates as the following criterion to be minimized (more details about the derivation are available in Appendix A):
\begin{equation}
\begin{split}
&\: L_{\mathit{CROSS}}(\bm{F_t}) \\
= &\sum_{\bm{P}} \mathit{min}(1+\frac{\partial F_{t}^x(\bm{P})}{\partial x}, 0)^2 + \mathit{min}(1+\frac{\partial F_{t}^y(\bm{P})}{\partial y}, 0)^2
\end{split}
\end{equation}

Moreover, we further encourage the flow to preserve the segmentation masks of cardiac structures $S\in$ \{LVC, LVM, RVC\}. The warped segmentation masks of these structures should approximately match the ground-truth masks on the corresponding frame. Let us note $M_{\mathit{ED}}^{S}$ and $M_{\mathit{ES}}^{S}$ the binary ground-truth segmentation mask (of pixel intensity value 0 or 1) of $S$ at the instants of ED and ES (the only instants for which the ground-truth is available in the ACDC training set). This constraint on the flow between ED and ES is based on the Dice coefficient
\begin{equation}
L_{\mathit{GT}}(\bm{F_{\mathit{ES}}}) = \sum_{S \in \mathit{\{LVC, LVM, RVC\}}}Dice(M_{\mathit{ED}}^{S}, M_{\mathit{ES}}^{S}\circ W_{\bm{F_{\mathit{ES}}}})  
\end{equation}
The formula of the $Dice$ function is provided in Appendix A.

Finally, the overall loss function for training the ApparentFlow-net is a linear combination of the terms $L_{\mathit{IMG}}$, $L_{\mathit{CROSS}}$ and potentially $L_{\mathit{GT}}$. We adopt a semi-supervised approach for which $L_{\mathit{GT}}$ is applied when ground-truth segmentation is available:
\begin{equation}
L_{\mathit{flow}}(\bm{F_t}) = L_{\mathit{IMG}}(\bm{F_t}) + p_{1}L_{\mathit{CROSS}}(\bm{F_t}) + p_{2}\bm{1}_{\mathit{t=ES}}L_{\mathit{GT}}(\bm{F_t})
\end{equation}
where $\bm{1}_{\mathit{t=ES}}$ is the indicator function for the event $t=$ ES. $\bm{1}_{\mathit{t=ES}}$ is necessary as for the instants $t$ other than ED and ES, the ground-truth segmentation is not provided in ACDC. Please note that this is a typical method of semi-supervised learning. It makes use of a small amount of labeled data (the images with ground-truth segmentation) and a large amount of unlabeled data (the images without ground-truth).

\begin{figure*}[]
\centering
\includegraphics[width=17cm, height=4.2cm]{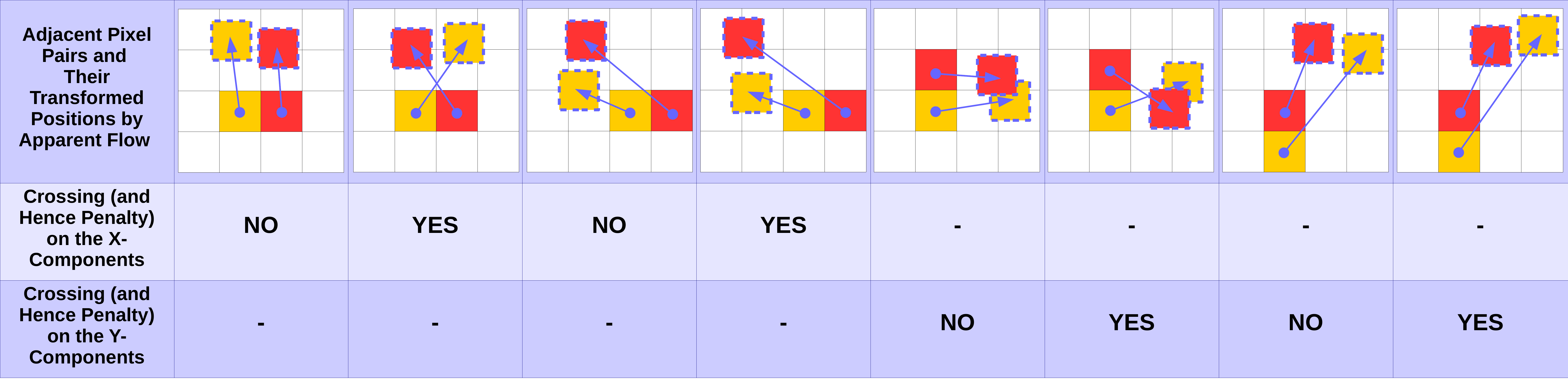}
\caption{Examples of adjacent pixel pairs transformed by apparent flow for which the crossing penalty applies or not.}
\end{figure*}

\subsection{Feature Extraction Step 2: Segmentation}
In this step, an existing model for segmentation proposed in \cite{Zheng:2018}, the LVRV-net, is applied to segment MRI image stacks as presented in \cite{Zheng:2018}. With the concept of propagation along the long axis, this method was proven to be robust, as the results achieved on several different datasets are all comparable or even better than the state-of-the-art. For more details about the structure, training and application of the LVRV-net, please refer to \cite{Zheng:2018}. When we train and evaluate our method on the ACDC training set (100 cases), in each fold of a 5-fold cross-validation, the trained LVRV-net as given by \cite{Zheng:2018} is finetuned with the 80 cases used for training before being applied on the remaining 20 cases; for the evaluation of our method on ACDC testing set (50 cases), the trained LVRV-net is first finetuned with the 100 cases of ACDC training set.

In fact, in \cite{Zheng:2018}, LVRV-net was trained to start the segmentation propagation from a given slice on which the ventricle cavities are supposed to be present. In other words, it was only trained to identify LV and RV labels on the slices below the base. So it might not work well if the basal slice is not determined in a stack and if the top slice in the volumetric image is located above the base. In this case, if we apply the original LVRV-net starting from the top slice, it might make a false positive prediction. With finetuning on ACDC, we find that this issue is solved. In general, the finetuned LVRV-net successfully learns from the ground-truth segmentation masks of ACDC that no foreground pixel is present (i.e. predict everything to be background) on the slices above the base and start segmentation propagation only when the base is reached. So it is no longer necessary to determine the basal slice manually. On the resulting sets of segmentation masks, we can hence also determine the location of the base, which is necessary for the calculation of volumes and the determination of sub-stacks for motion extraction as we will present later.

With the segmentation mask, we determine $\bm{B_{L}}$ and $\bm{B_{R}}$, the barycenters of LVC and RVC respectively. Then all the pixels $\bm{P}$ labeled to LVM on the segmentation mask are divided into 6 segments, depending on in which interval $[k\pi/3, (k+1)\pi/3[$ for $k$ in $[0,5]$ the angle between the vectors $\bm{B_{L}P}$ and $\bm{B_{L}B_{R}}$ is. An example of the resulting 6 segments are shown in Fig.1. This division of segments is inspired by the 17-segment system recommended by the American Heart Association (AHA) in \cite{Cerqueira:2002}. Indeed, in the AHA system, on all short-axis slices around the base and at the level of mid-cavity, the myocardium is divided into 6 segments.

\subsection{Feature Extraction Step 3: Shape-Related Features}
Based on the segmentation masks generated in the previous step, we estimate the volumes of LVC, LVM and RVC of a case at ED and ES. For each of the two phases, the volume of LVC is calculated by approximating the LVC between two adjacent slices as a truncated cone and summing up all the truncated cone volumes:
\begin{equation}
\mathit{V_{LVC}} = \sum_{i} (S_i + S_{i+1} + \sqrt{S_iS_{i+1}})(L_{i+1}-L_{i})/3
\end{equation} 
where $S_i$ is the area of LVC on the slice $i$ and $L_i$ is the slice position along the long axis. The volume of LVM and RVC is calculated in a similar way. Then we normalize all the volumes by the corresponding body surface area (BSA) of the subject, which is a traditional practice based on the assumption that BSA is related to the metabolic rate. BSA can be computed from the height and the weight provided in ACDC (using the Mosteller formula BSA=$\sqrt{\mathit{height}*\mathit{weight}}/60$ ).

With the segmentation masks and volumes at ED and ES, we then compute the 7 shape-related features as listed in the first 7 rows of Table 1.

\begin{table}[]
\caption{The extracted features used by our classification model}
\centering
\begin{tabular}{c c c}
\hline
\noalign{\vskip 0.0in}
\multicolumn{1}{|c}{Feature} & \multicolumn{1}{|c|}{Notion (and Definition)} \\ 
\hline
\multicolumn{1}{|c}{RVC volume at ED} & \multicolumn{1}{|c|}{$V_\mathit{RVC,ED}$} \\ 
\hline
\multicolumn{1}{|c}{LVC volume at ES} & \multicolumn{1}{|c|}{$V_\mathit{LVC,ES}$} \\ 
\hline
\multicolumn{1}{|c}{RVC ejection fraction} & \multicolumn{1}{|c|}{$\mathit{EF_{RVC}}$ $(= 1-\mathit{V_{RVC,ES}/V_{RVC,ED})}$} \\ 
\hline
\multicolumn{1}{|c}{LVC ejection fraction} & \multicolumn{1}{|c|}{$\mathit{EF_{LVC}}$ $(= 1-\mathit{V_{LVC,ES}/V_{LVC,ED})}$} \\ 
\hline
\multicolumn{1}{|c}{Ratio between RVC and} & \multicolumn{1}{|c|}{$\mathit{R_{RVCLV,ED}}$} \\ 
\multicolumn{1}{|c}{LV volumes at ED} & \multicolumn{1}{|c|}{$(=\mathit{V_{RVC,ED}/(V_{LVC,ED}+V_{LVM,ED}))}$} \\ 
\hline
\multicolumn{1}{|c}{Ratio between LVM and} & \multicolumn{1}{|c|}{$\mathit{R_{LVMLVC,ED}}$} \\ 
\multicolumn{1}{|c}{LVC volumes at ED} & \multicolumn{1}{|c|}{$(=\mathit{V_{LVM,ED}/V_{LVC,ED})}$} \\ 
\hline
\multicolumn{1}{|c}{Maximal LVM thickness} & \multicolumn{1}{|c|}{$\mathit{MT_{LVM,ED}}$} \\ 
\multicolumn{1}{|c}{in all the slices at ED} & \multicolumn{1}{|c|}{} \\ 
\hline
\multicolumn{1}{|c}{Radius motion} & \multicolumn{1}{|c|}{$\mathit{RMD}$} \\ 
\multicolumn{1}{|c}{disparity} & \multicolumn{1}{|c|}{} \\ 
\hline
\multicolumn{1}{|c}{Thickness motion} & \multicolumn{1}{|c|}{$\mathit{TMD}$} \\ 
\multicolumn{1}{|c}{disparity} & \multicolumn{1}{|c|}{} \\ 
\hline
\end{tabular}
\end{table}

\begin{figure*}[]
\centering
\includegraphics[width=17.5cm, height=6.8cm]{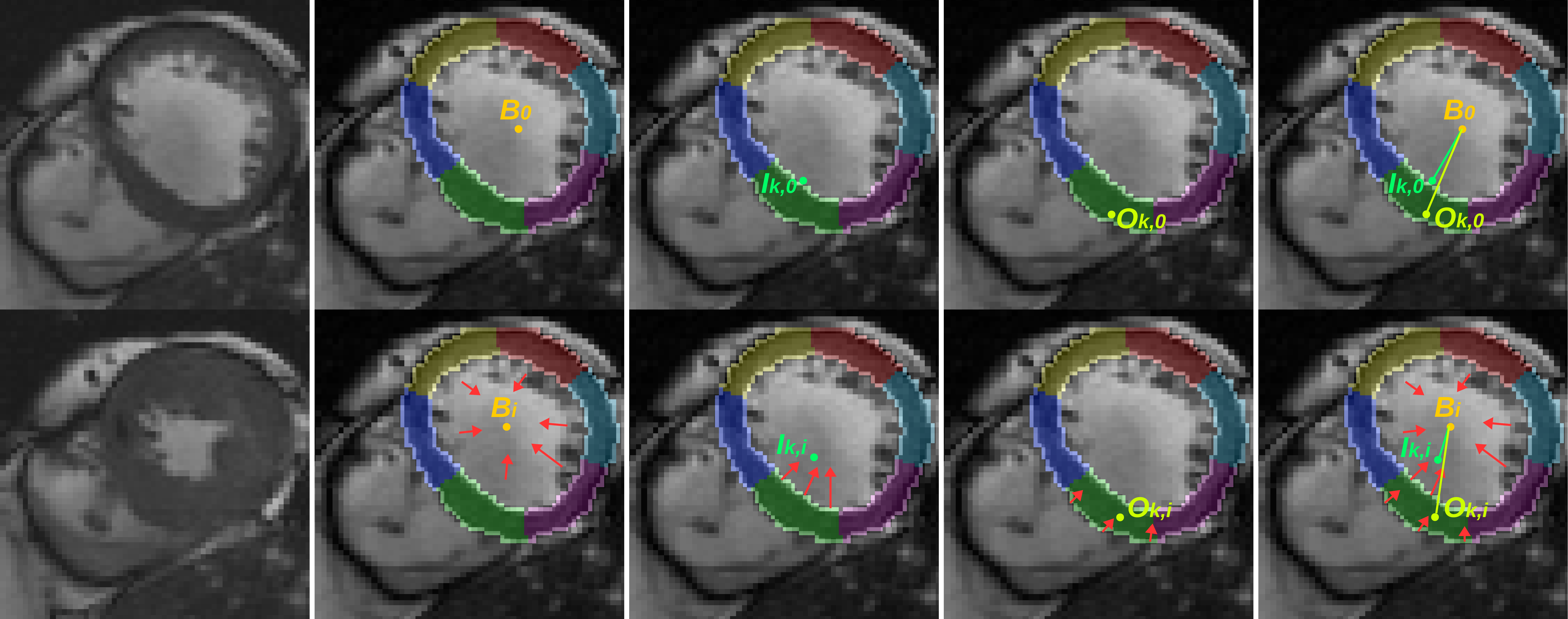}
\caption{Definitions of $\bm{B_i}$, $\bm{I_{k,i}}$, $\bm{O_{k,i}}$, $RA_{k,i}$ and $T_{k,i}$, for the extraction of motion-characteristic time series. The first row shows the definitions at $t_0$; the second row presents the definitions at $t_i$ for $i \in [1,9]$. \quad\quad 1st column: Frames at $t_0$ and $t_i$, based on which the apparent flow is generated.  \quad\quad 2nd column: $\bm{B_i}$ is the barycenter of warped LVC (segmented at $t_0$) at $t_i$. \quad\quad 3rd column: $\bm{I_{k,i}}$ is the barycenter of the warped inner boundary of segment $S_k$ at $t_i$. \quad\quad 4th column: $\bm{O_{k,i}}$ is the barycenter of the warped outer boundary of segment $S_k$ at $t_i$. \quad\quad 5th column: $RA_{k,i} = |\bm{B_iI_{k,i}}|/\mathit{BSA}$, $T_{k,i} = |\bm{B_iO_{k,i}}|/\mathit{BSA} - RA_{k,i}$ .}
\end{figure*}

\begin{figure*}[]
\centering
\includegraphics[width=18cm, height=19cm]{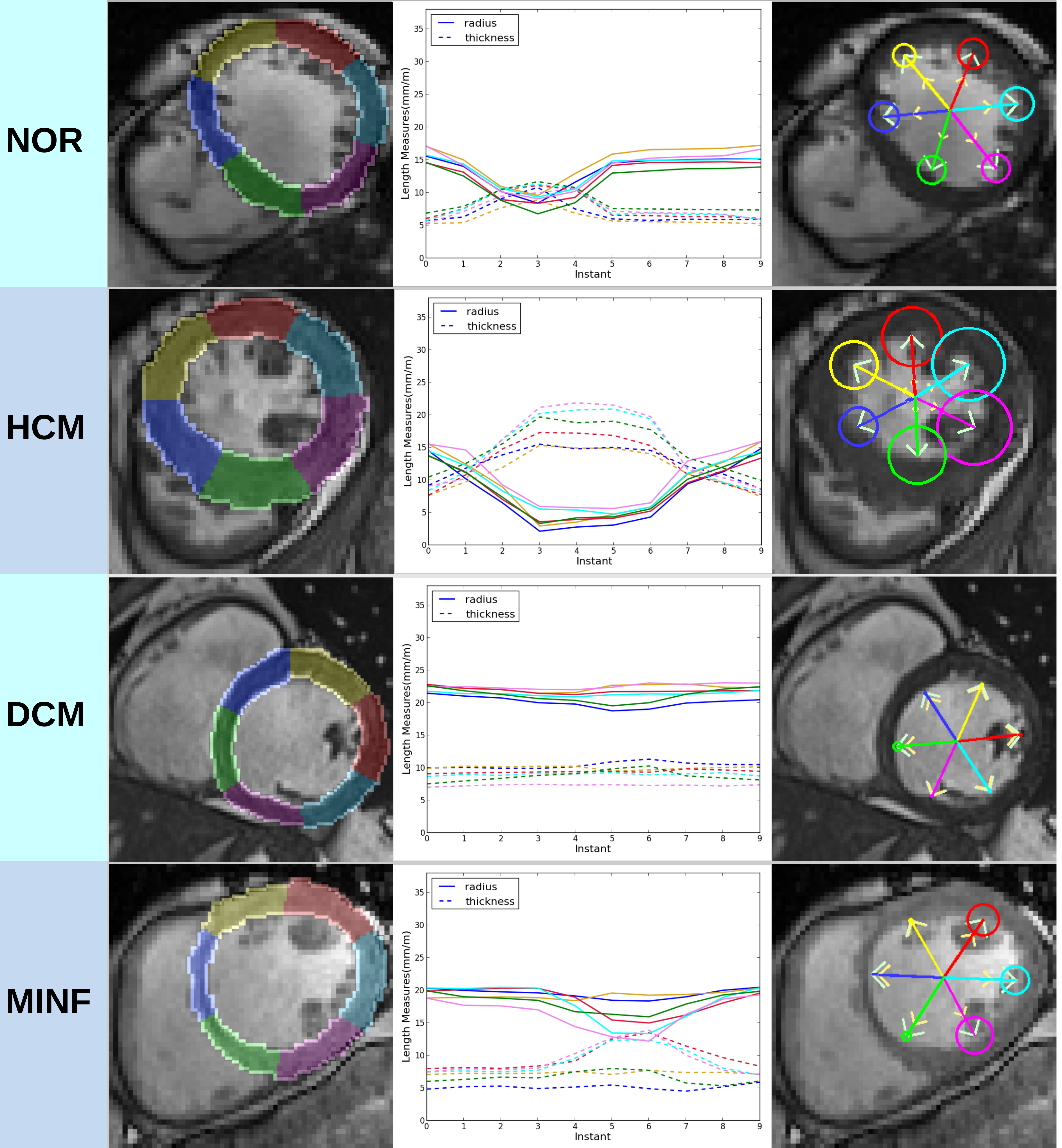}
\caption{Examples of typical slice from 4 of the 5 pathological categories in ACDC. \quad\quad 1st column: the segmentation of the 6 myocardial segments (the boundaries of the segmentation masks are marked by lighter colors). \quad\quad 2nd column: time series of the radius (solid lines) and the thickness (dotted lines) of the 6 segments. \quad\quad 3rd column: a visualization of the motion information. For each segment, the radius connecting the LVC barycenter $\bm{B_0}$ and the segment inner boundary barycenter $\bm{I_{k,0}}$ (marked by the light green arrow) at ED is plotted. The segment inner boundary barycenter at ES is marked by the light orange arrow. The radius of the circle is proportional to the difference of segment thicknesses at ES and ED ($\Delta T_{k,i}$).}
\end{figure*}

\subsection{Feature Extraction Step 4: Motion-Characteristic Features}

\subsubsection{Slice Selection}
For each case, let $S$ be the image stack at ED (following the Notation part in the previous section). Given the segmentation masks of each slice generated in Step 2, we note $i_1$ the index of the first slice on which RVC mask is present (roughly the first slice below the base), and $i_2$ the index of the last slice on which LVC mask is present (roughly the last slice above the apex), and $h=i_2-i_1+1$. Then we focus on extracting motion information from the sub-stack $\mathit{S_{mid}} = S[i_1+0.1h, i_2+1-0.2h]$. Please note that among the slices between the base and the apex, we exclude the top 10\% and the bottom 20\% and consider the remaining 70\% in the middle, since the out-of-plane motion is particularly large in the slices close to the base or the apex.

\subsubsection{Frame Sampling}
As presented in Fig.1, for each slice in $\mathit{S_{mid}}$, let us note $f$ the number of frames available (all the frames together form a cardiac cycle). We sample 10 frames of instant $t_i$ for $i$ in [0,9], such that $t_0$ is the instant of ED and $t_i=\mathit{round}(t_0+i*f/10) \mod f$ for $i$ in [1,9]. The 10 sampled frames hence cover the whole cardiac cycle. Applying the ApparentFlow-net of Step 1 in all the 9 pairs of frame $(t_0,t_i)$, we obtain 9 apparent flow maps $\bm{F_{\mathit{t_i}}}$. Hence for each pixel $\bm{P}$, we get its warped position $W_{\bm{F_{\mathit{t_i}}}}(\bm{P})$ for $i$ in [1,9].

\subsubsection{Time Series Extraction}
Then, with the convention that $F_{\mathit{t_0}}$ is the null apparent flow (and hence $W_{\bm{F_{\mathit{t_0}}}}$ is the identity function), the barycenter of LVC at $t_i$ for $i \in [0,9]$, $\bm{B_i}$, is defined as the average of $W_{\bm{F_{\mathit{t_i}}}}(\bm{P})$ for all the pixels $\bm{P}$ labeled as LVC on the segmentation mask $\mathit{M_{ED}}$ at $t_0$ (the 2nd column of Fig.5):
\begin{equation}
\bm{B_i} = average(\{W_{\bm{F_{\mathit{t_i}}}}(\bm{P}) \: | \: \bm{P} \in LVC \: on \: \mathit{M_{ED}} \})
\end{equation} 
In a similar way, for each myocardial segment $S_k$ ($k \in [0,5]$) and each instant $t_i$ ($i \in [0,9]$), we define $\bm{I_{k,i}}$, the barycenter of the inner boundary of the myocardial segment $S_k$ at $t_i$ (the 3rd column of Fig.5):

\begin{equation}
\begin{split}
\bm{I_{k,i}} = & \: average(\{W_{\bm{F_{\mathit{t_i}}}}(\bm{P}) \: | \: \bm{P} \in LVC \: on \: \mathit{M_{ED}} \\
& \: \& \: \bm{P} \: has \: neighboring \: pixel(s) \in S_k\})
\end{split}
\end{equation} 
and $\bm{O_{k,i}}$, the barycenter of the outer boundary of the myocardial segment $S_k$ at $t_i$ (the 4th column of Fig.5):
\begin{equation}
\begin{split}
\bm{O_{k,i}} = & \: average(\{W_{\bm{F_{\mathit{t_i}}}}(\bm{P}) \: | \: \bm{P} \in S_k  \\
& \& \: \bm{P} \: has \: neighboring \: pixel(s) \in background \: on \: \mathit{M_{ED}}\})
\end{split}
\end{equation} 
Finally, as shown in the 4th column of Fig.5, we define the radius of $S_k$ at $t_i$ normalized by BSA: 
\begin{equation}
RA_{k,i} = |\bm{B_iI_{k,i}}|/\mathit{BSA}
\end{equation}
and the thickness of $S_k$ at $t_i$ normalized by BSA:
\begin{equation}
T_{k,i} = |\bm{B_iO_{k,i}}|/\mathit{BSA} - RA_{k,i}
\end{equation}
We hence generate two time series $\{RA_{k,i}:i\in[0,9]\}$ and $\{T_{k,i}:i\in[0,9]\}$ to represent the contraction and the thickening of $S_k$.

\subsubsection{Visual Correspondence between Time Series and Pathologies}
We compute the two time series introduced above for all the slices in $S_{\mathit{mid}}$ of all the cases in ACDC. From the majority of the cases, we manage to visually identify several typical slices with the time series characterizing the motion of the corresponding category. Examples of such typical slices are presented in Fig.6. %(no example of RVA is shown as the abnormality of an RVA case might not be present on LV and hence not on the two time series)
To sum up our observation on the typical slices of each category as shown in Fig.6:\\
\textbullet \ NOR: all segments have similar radius and thickness at all instants; their contraction and thickening are synchronous and with comparable scales. \\
\textbullet \ HCM: the segments not only look proportionally thicker at ED, but also thicken more %(more increment on the dotted lines on the graph in the second column, larger circles on the image in the third column) 
and contract stronger %(more decrease on the solid lines on the graph in the second column, proportionally less distance between the LVC barycenter and the light orange arrows on the image in the third column)
in the radial direction. \\
\textbullet \ DCM: the radiuses are quite large; the segments are moving so little that neither contraction nor thickening is obvious%(flat lines in the graph, barely visible circles, the light orange arrows almost overlap the light green arrows on the image in the third column)
. \\
\textbullet \ MINF: the radiuses are quite large; some segments are clearly much more active than others%(disparity around ES among the solid lines and the dotted lines in the graph, big circles versus tiny circles as well as close pairs versus separated pairs of light green and light orange arrows on the image in the third column)
.

\subsubsection{Motion-Characteristic Feature Values}
To better distinguish DCM and MINF cases, we define two additional feature values which often indicate the abnormal contraction described in the definition of MINF.

The first one is ``radius motion disparity'' (RMD). Given a case, we consider the set of radius series $\{RA_{k,i}:i\in[0,9]\}$ of all the segments $S_k$ on all the slices in the sub-stack $S_{\mathit{mid}}$ (e.g. if there are 4 slices in $S_{\mathit{mid}}$, we consider a set of $6\times4=24$ time series). We first define the disparity of motion over all the segments in $S_{\mathit{mid}}$ at the instant $t_i$ as the difference between the maximum and minimum contraction at $t_i$:
\begin{equation}
\mathit{RD_i} = \mathit{\max_{S_k \in S_{\mathit{mid}}}} RA_{k,i}/RA_{k,0} - \mathit{\min_{S_k \in S_{\mathit{mid}}}} RA_{k,i}/RA_{k,0} 
\end{equation}
Then RMD is defined as the maximum disparity along the cardiac cycle:
\begin{equation}
\mathit{RMD} = \mathit{\max_{i \in [0,9]}} \mathit{RD_i}
\end{equation}

The second motion-characteristic feature value is named ``thickness motion disparity'' (TMD). For each slice $s$ in $S_{\mathit{mid}}$ and each $t_i$, we define the thickness motion disparity of the slice $s$ at $t_i$ as
\begin{equation}
\mathit{TD_{s,i}} = \: (\mathit{\max_{k \in [0,5]}} T_{k,i} - \mathit{\min_{k \in [0,5]}} T_{k,i} ) / \mathit{\min_{k \in [0,5]}} T_{k,0}
\end{equation}
where we normalize the thicknesses by the minimum segment thickness at $t_0$ on slice $s$ taking into account that myocardial thickness may vary across slice.

Finally, TMD is defined as
\begin{equation}
\mathit{TMD} = \mathit{\max_{s \in S_{\mathit{mid}}, \: i \in [0,9]}} \mathit{TD_{s,i}} 
\end{equation}

\subsection{Classification}

\subsubsection{4-Classifier Classification Model}
Using the 7 shape-related features and the 2 motion-characteristic features as input, a classification model is proposed (Fig.2) to classify the 5 pathological categories of ACDC. It consists of 4 binary classifiers:\\
\textbullet \ RVA classifier: RVA cases v.s. all the other cases.\\
\textbullet \ HCM classifier: HCM cases v.s. MINF, DCM and NOR cases.\\
\textbullet \ DCM classifier: DCM cases v.s. MINF and NOR cases.\\
\textbullet \ MINF classifier: MINF cases v.s. NOR cases.\\

The 4 binary classifications are arranged in increasing order of difficulty of the binary classification tasks. RVA and HCM cases can be identified based on the commonly used shape-related features. So they are classified first. DCM and MINF cases are somewhat similar in terms of sizes and ejection fractions. We use the novel motion-characteristic features to better distinguish them. Hence this more difficult classification is performed at the end.

\subsubsection{Explainable Manual Feature Selection}
To keep the classifiers simple, limit their risk of overfitting and increase their explainability, we chose no more than 3 features for each classifier as shown in Table 2:\\
\textbullet \ For RVA classifier, according to the definition provided by ACDC, the RVC volume at ED and the RVC ejection fraction are the most relevant features. We add one more feature, the ratio between RVC and LV volumes at ED, as we find that most RVA cases have disproportionately large RVC.\\
\textbullet \ For HCM classifier, LVC ejection fraction and maximal LVM thickness are selected according to the definition of HCM. The ratio between LVM and LVC volumes at ED is added because with most HCM cases this ratio is exceptionally high due to the exceptional myocardial thickness .\\
\textbullet \ For DCM classifier, as DCM cases are usually dilated at ED and inactive from ED to ES, their volumes of LVC at ES can be exceptionally large. So this feature is used. In addition, we also use radius motion disparity and thickness motion disparity. \\
\textbullet \ For MINF classifier, by definition, LVC ejection fraction is enough to distinguish MINF cases from NOR cases

\subsubsection{Model of Classifiers}
Each of the 4 classifiers is just a ridge logistic regression model. For a training case of index $m$, if we note $f_{m,i}$ the $i$-th feature values used as input of the classifier and $y_m$ (-1 or 1, corresponding to no or yes) the binary ground-truth of the case, then the classifier is trained by minimizing
\begin{equation}
\begin{split}
&L_{\mathit{classifier}}(\{p_i\}, b) \\
= \:&\frac{1}{2}\sum_{i}p_i^2 + C\sum_{m}\mathit{log}\Big(\mathit{exp}\big(-y_m(\sum_{i}p_if_{m,i} + b)\big)+1\Big) 
\end{split}
\end{equation}
with respect to the parameters $\{p_i\}$ and $b$. $C$ is the inverse of regularization strength. After the training is done, given a case of index $l$ and feature values $f_{l,i}$, the prediction the sign of $\sum_{i}p_if_{l,i} + b$. If it is non-negative, the prediction of the trained classifier is yes; otherwise it is no.

\subsubsection{Flexibility and Versatility of the Model}
Finally, we would also like to point out that the 4 classifiers function independently. While they are grouped together to form the proposed classification model in this paper, they can certainly be applied separately or grouped in a different manner in other situations if appropriate. This proposed classification model is hence very flexible and versatile.

\begin{table}[]
\caption{The input features of the 4 binary classifiers}
\centering
\begin{tabular}{c c}
\hline
\noalign{\vskip 0.0in}
\multicolumn{1}{|c}{} & \multicolumn{1}{|c|}{Input Feature(s)} \\ 
\hline
\multicolumn{1}{|c}{RVA Classifier} & \multicolumn{1}{|c|}{$\mathit{V_{RVC,ED}}$, $\mathit{EF_{RVC}}$, $\mathit{R_{RVCLV,ED}}$} \\ 
\hline
\multicolumn{1}{|c}{HCM Classifier} & \multicolumn{1}{|c|}{$\mathit{EF_{LVC}}$, $\mathit{R_{LVMLVC,ED}}$, $M\mathit{T_{LVM,ED}}$} \\ 
\hline
\multicolumn{1}{|c}{DCM Classifier} & \multicolumn{1}{|c|}{$\mathit{V_{LVC,ES}}$, $\mathit{RMD}$, $\mathit{TMD}$} \\ 
\hline
\multicolumn{1}{|c}{MINF Classifier} & \multicolumn{1}{|c|}{$\mathit{EF_{LVC}}$} \\ 
\hline
\end{tabular}
\end{table}

\section{Experiments and Results}
We evaluate our method in two different ways. On the one hand, the model is trained with ACDC training set and then tested on ACDC testing set. On the other hand, a 5-fold cross-validation is performed on ACDC training set. For the latter, the 100 cases of ACDC training set are partitioned into 5 subsets of 20 cases, such that in each subset there are exactly 4 cases of each of the 5 categories. 

In addition, we also analyze the proposed model by comparing it with various other models. Since the ground-truth category is only available for the cases in the training set (and not for those in the testing set), this analysis is based on the results on the training set.

\subsection{Training ApparentFlow-net}

\subsubsection{Parameters and Data}
In the training process with the whole ACDC training set, as well as in each of the 5 training processes of the 5-fold cross validation, the ApparentFlow-net is trained using the loss function $L_{\mathit{flow}}(\bm{F_t})$ introduced in the Method section for 50 epochs with batch size 16. In terms of loss function parameter, we empirically find that $p_1=10^3$ and $p_2=10^5$ work well. These values are hence used for training. In terms of training data, for each case in the corresponding training set, we use the slices in the sub-stack $S[i_1+0.2h, i_2+1-0.2h]$ (with the notation introduced in the sub-section 3.5.1). In other words, we approximately exclude the top 20\% and the bottom 20\% of all the slices covering the LV cavity, and select the remaining 60\% in the middle. This slice selection for training (middle 60\%) is slightly more conservative than that for the application of the method (middle 70\%). This design is aimed to further reduce the impact of the out-of-plane motion in training. For each selected slice, the frame pairs of indices (ED, $t$) for all frame index $t$ are used to train the ApparentFlow-net. Only when $t=$ ES, the term $L_{\mathit{GT}}(\bm{F_t})$ in $L_{\mathit{flow}}(\bm{F_t})$ using the segmentation ground truth is applied. With our automatic slice selection approach, in total, there are 13672 frame pairs used for training in the ACDC training set. Among the 13672 frame pairs, only 515 pairs (3.77\%) come with segmentation ground-truth such that the term $L_{\mathit{GT}}(\bm{F_t})$ applies.

\subsubsection{Performance}
To measure its performance, in each evaluation of the 5-fold cross-validation, for all the slices in the sub-stack $S[i_1+0.2h, i_2+1-0.2h]$ of all the 20 cases for evaluation, we apply the trained ApparentFlow-net to generate $\bm{F_{ES}}$. Then we use it to warp the ground-truth segmentation mask at ES, noted as  $M_{\mathit{ES}}$, to obtain $M_{\mathit{ES}}\circ W_{\bm{F_{\mathit{ES}}}}$. $M_{\mathit{ES}}\circ W_{\bm{F_{\mathit{ES}}}}$ is then compared with $M_{\mathit{ED}}$, the corresponding ground-truth masks at ED, using Dice coefficient (2D version) on LVM, LVC and RVC. Overall, the means(standard deviations) of Dice coefficients achieved on LVM, LVC and RVC in the 5-fold cross-validation are reported in Table 3.

\begin{table}[t]
\caption{The mean(standard deviation) of Dice coefficients achieved by comparing $M_{\mathit{ES}}\circ W_{\bm{F_{\mathit{ES}}}}$ and $M_{\mathit{ED}}$ for 3 cardiac structures in the 5-fold cross-validation on ACDC training set.}
\centering
\begin{tabular}{c ccc}
\hline
\noalign{\vskip 0.0in}
\multicolumn{1}{|c}{Training Method} & \multicolumn{1}{|c}{} & \multicolumn{1}{c}{Dice} & \multicolumn{1}{c|}{} \\ 
\cline{2-4}
%\hline
\multicolumn{1}{|c}{} & \multicolumn{1}{|c}{LVM} & \multicolumn{1}{|c}{LVC} & \multicolumn{1}{|c|}{RVC} \\ 
\hline
\multicolumn{1}{|c}{semi-supervised} & \multicolumn{1}{|c}{0.84(0.07)} & \multicolumn{1}{|c}{0.94(0.07)} & \multicolumn{1}{|c|}{0.87(0.19)} \\ 
\multicolumn{1}{|c}{(proposed)} & \multicolumn{1}{|c}{} & \multicolumn{1}{|c}{} & \multicolumn{1}{|c|}{} \\ 
\hline
\multicolumn{1}{|c}{unsupervised} & \multicolumn{1}{|c}{0.76(0.08)} & \multicolumn{1}{|c}{0.93(0.06)} & \multicolumn{1}{|c|}{0.83(0.22)} \\ 
\hline
\end{tabular}
\end{table}

Additionally, we also visually evaluate the apparent flow generated by the ApparentFlow-net. We find that the apparent flow is indeed good enough to characterize the cardiac motion of the typical cases in the pathological categories. Several examples are given in  Appendix D.

\subsubsection{Importance of Supervision in Training}
In order to understand the importance of the small amount of segmentation ground-truth used in the proposed semi-supervised learning method, we also train a variant of ApparentFlow-net using only unsupervised learning. The only modification is the removal of the term $L_{\mathit{GT}}(\bm{F_t})$ from $L_{\mathit{flow}}(\bm{F_t})$ such that the variant is trained without any ground-truth for supervision. As reported in Table 3, the means of Dice coefficients on LVM, LVC and RVC are all lower than the corresponding results achieved by the semi-supervised learning method. In particular, there is a large drop from 0.84 to 0.76 on the mean of Dice coefficient on LVM. So the proposed semi-supervised learning method is indeed better than its unsupervised learning variant by making efficient use of the small amount of segmented images.

\subsection{Finetuning LVRV-net}

LVRV-net is already trained in \cite{Zheng:2018} for 80 epochs on a subset of about 3000 cases of UK Biobank (\cite{Petersen:2016}). In the training process with the whole ACDC training set, as well as in each of the 5 training processes of the 5-fold cross validation, LVRV-net is finetuned for 920 epochs on the corresponding training data, with exactly the same loss function and training parameters as given in \cite{Zheng:2018}. With the finetuning, the means (standard deviations) of 3D Dice coefficients achieved on LVC, LVM and RVC segmentation volumes in the 5-fold cross-validation are 0.94(0.06), 0.90(0.03) and 0.89(0.12).

\subsection{Proposed Classification Model}
Apparent flows and segmentation masks are generated by the ApparentFlow-net and the finetuned LVRV-net, from which the 7 shape-related features and the 2 motion-characteristic features are extracted. Then the 4 ridge logistic regression binary classifiers are implemented using Scikit-learn \cite{Pedregosa:2011} and trained on the cases of the categories they are supposed to classify. For example, DCM classifier is trained on the cases of NOR, MINF and DCM; the cases of RVA or HCM are not used to train it. In terms of classifier parameter, we empirically find that $C=50$ works well and use it in this paper. The performances of some variants with different values of $C$ are provided in Appendix B.

\begin{table*}[]
\caption{The classification performance on the testing set (50 cases) and training set (100 cases) of ACDC by different models}
\centering
\begin{tabular}{c ccc}
\hline
\noalign{\vskip 0.0in}
\multicolumn{1}{|c}{Model} & \multicolumn{1}{|c}{Testing Set}  & \multicolumn{1}{|c}{Training Set} & \multicolumn{1}{|c|}{Evaluation Method on Training Set} \\
\multicolumn{1}{|c}{} & \multicolumn{1}{|c}{Accuracy} & \multicolumn{1}{|c}{Accuracy} & \multicolumn{1}{|c|}{} \\  
\hline
\multicolumn{1}{|c}{proposed model} & \multicolumn{1}{|c}{94\%} & \multicolumn{1}{|c}{95\%} & \multicolumn{1}{|c|}{5-fold cross-validation}  \\  
\hline
\multicolumn{1}{|c}{\cite{Isensee:2017}} & \multicolumn{1}{|c}{92\%} & \multicolumn{1}{|c}{94\%} & \multicolumn{1}{|c|}{5-fold cross-validation} \\ 
\hline
\multicolumn{1}{|c}{\cite{Wolterink:2017}} & \multicolumn{1}{|c}{86\%} & \multicolumn{1}{|c}{91\%} & \multicolumn{1}{|c|}{4-fold cross-validation} \\ 
\hline
\multicolumn{1}{|c}{\cite{Cetin:2017}} & \multicolumn{1}{|c}{92\%} & \multicolumn{1}{|c}{100\%} & \multicolumn{1}{|c|}{forward feature selection and leave-one-out cross-validation} \\ 
\hline
\multicolumn{1}{|c}{\cite{Khened:2017}} & \multicolumn{1}{|c}{96\%} & \multicolumn{1}{|c}{90\%} & \multicolumn{1}{|c|}{70 cases for training, 20 for validation, 10 for evaluation}  \\
\hline
\multicolumn{1}{|c}{\cite{Khened:2018}} & \multicolumn{1}{|c}{\textbf{100\%}} & \multicolumn{1}{|c}{N.A.} & \multicolumn{1}{|c|}{N.A.}  \\

\hline

\end{tabular}
\end{table*}

\begin{figure}[]
\centering
\includegraphics[width=8.0cm, height=3.2cm]{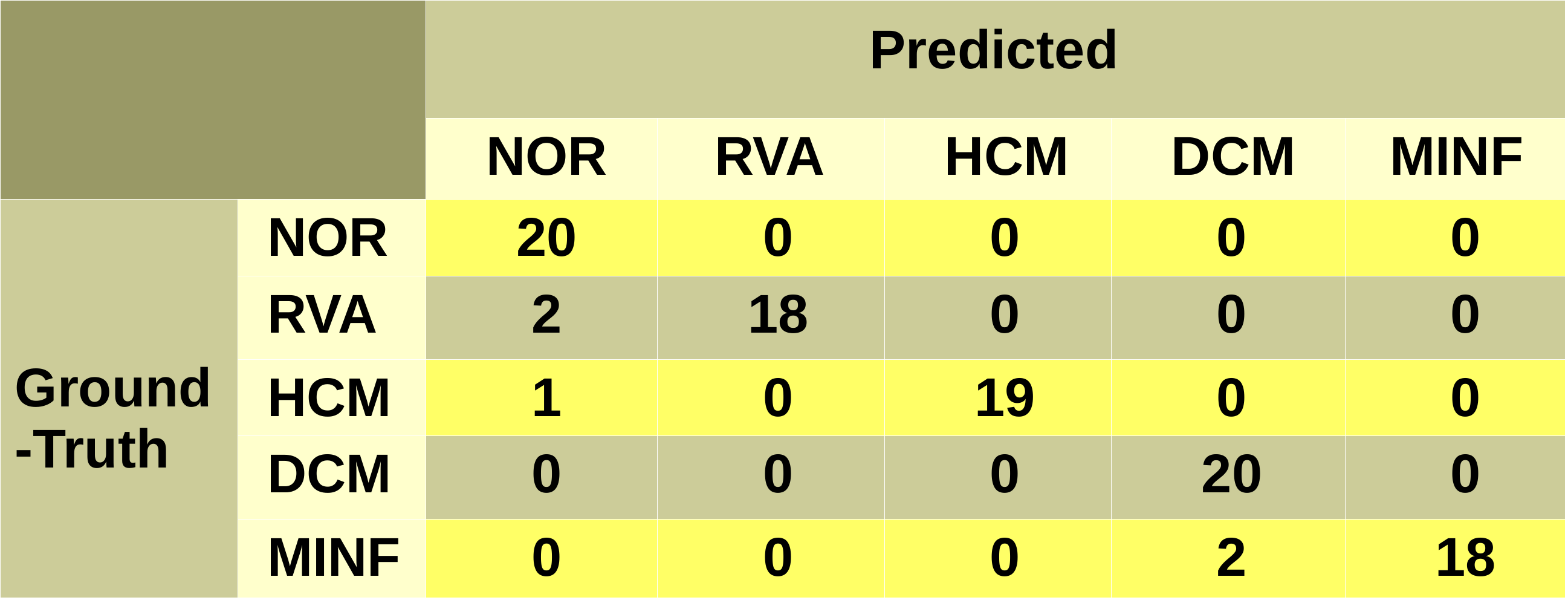}
\caption{The confusion matrix of the predictions by the proposed classification model in the 5-fold cross-validation on the training set of ACDC.}
\end{figure}

\subsubsection{Classification Performance}
As presented in Table 4, on the testing set, the accuracy of our model is 94\%. In the 5-fold cross-validation on the training set, our method achieves an accuracy of 95\%. Hence our model achieves performances that are comparable to those of the state-of-the-art on both the training set and the testing set. This is quite remarkable because, in contrast to the state-of-the-art, each classifier in our model uses only up to three features and has only up to 4 parameters. In total, our model uses 9 features and has 14 parameters. And each feature is selected in a clearly explainable manner. On the testing set, among the two methods with performances better than ours, \cite{Khened:2017} uses a random forest of 100 trees and \cite{Khened:2018} applies a more sophisticated ensemble system. Therefore, those classification models are less straightforward to interpret than ours. Furthermore, since our model has very similar performances on the training and testing sets, there seems to be little overfit.

Based on the confusion matrix of the prediction in the 5-fold cross-validation on the ACDC training set (Fig.7), for the binary classification of NOR, RVA, HCM, DCM and MINF, we calculate and find that the precision values are 0.87, 1.00, 1.00, 0.91 and 1.00; the recall values are 1.00, 0.90, 0.95, 1.00 and 0.90.

\subsubsection{Interpretation of Mis-Classification}
As our classifier can be interpreted easily, we figure out for each of the 5 misclassified cases (Fig.7) why the prediction is different from the ground-truth. In fact, they all seem to be somewhat ambiguous in terms of pathological category:\\
\textbullet \ Patients 082 and 088 are of ground-truth RVA but are classified as NOR. According to our segmentation, they have $\mathit{V_{RVC,ED}}$, $\mathit{EF_{RVC}}$ and $\mathit{R_{RVCLV,ED}}$ values all very similar to that of the NOR cases. For instance, they have the third and the first lowest $\mathit{R_{RVCLV,ED}}$ values (0.755 and 0.691) among all the RVA cases, which are well in the range of that of the NOR cases. \\
\textbullet \ Patient 022 is of ground-truth HCM but is predicted as NOR. Unlike all the other HCM cases, patient 022 has both $\mathit{EF_{LVC}}$ (0.622) and $\mathit{MT_{LVM,ED}}$ (14.7mm) in the normal ranges, which makes it look like a NOR case. \\
\textbullet \ Patients 050 and 060 are of ground-truth MINF but are predicted as DCM. Their values of $\mathit{V_{LVC,ES}}$ (118.0$\mathit{mL/m^2}$ and 83.5$\mathit{mL/m^2}$) are the two highest among all the non-DCM cases and well in the range of that of the DCM cases. In terms of motion disparity, on $\mathit{RMD}$ and $\mathit{TMD}$, unlike the majority of the MINF cases, their values (0.245 and 1.173 for patient 050, 0.316 and 1.246 for patient 060) are also in the ranges of that of the DCM cases. For these reasons, the DCM classifier predicts them to be DCM cases.

\begin{table}[]
\caption{The parameters of the 4 logistic regression binary classifiers trained on the training set of ACDC}
\centering
\begin{tabular}{c c}
\hline
\noalign{\vskip 0.0in}
\multicolumn{1}{|c}{} & \multicolumn{1}{|c|}{Parameters of the Trained Classifier} \\ 
\hline
\multicolumn{1}{|c}{RVA Classifier} & \multicolumn{1}{|c|}{$0.010\mathit{V_{RVC,ED}} - 4.695\mathit{EF_{RVC}}$} \\ 
\multicolumn{1}{|c}{} & \multicolumn{1}{|c|}{$+ 14.012\mathit{R_{RVCLV,ED}} - 9.906$} \\ 
\hline
\multicolumn{1}{|c}{HCM Classifier} & \multicolumn{1}{|c|}{$8.434\mathit{EF_{LVC}}+4.614\mathit{R_{LVMLVC,ED}}$} \\ 
\multicolumn{1}{|c}{} & \multicolumn{1}{|c|}{$+0.420\mathit{MT_{LVM,ED}}-16.580$} \\ 
\hline
\multicolumn{1}{|c}{DCM Classifier} & \multicolumn{1}{|c|}{$0.104\mathit{V_{LVC,ES}}-0.918\mathit{RMD}$} \\ 
\multicolumn{1}{|c}{} & \multicolumn{1}{|c|}{$-7.758\mathit{TMD}-0.321$} \\
\hline
\multicolumn{1}{|c}{MINF Classifier} & \multicolumn{1}{|c|}{$-17.122\mathit{EF_{LVC}}+7.994$} \\ 
\hline

\end{tabular}
\end{table}

\subsubsection{Explaining the Classifiers}
The 4 binary classifiers are just logistic regression models. As presented in the previous section, their prediction depends on the sign of the sum $\sum_{i}p_if_{l,i} + b$. To understand what is learned from the data by the trained classifiers, in Table 5 we show the coefficients of the classifiers trained with all the relevant cases in ACDC. We find that the signs of the parameters $p_i$ all correspond to the positive or negative correlation between the feature and the binary classification task. For instance, in the trained RVA classifier, the signs of the coefficients of $\mathit{V_{RVC,ED}}$ and $\mathit{R_{RVCLV,ED}}$ are both positive, as a large RVC volume and a high ratio between the RVC and LV volumes are both indicators of RV abnormality; on the other hand, since low RVC ejection fraction usually signifies RV abnormality, the coefficient of $\mathit{EF_{RVC}}$ is negative. Similarly, such a correspondence applies to all the coefficients of the 3 other trained classifiers. In particular, for MINF classifier, the learned threshold on $\mathit{EF_{LVC}}$ to distinguish MINF cases from NOR cases is $7.994/17.122=0.467$, which can well separate them according to their definitions.

\subsection{Variants of the Proposed Classification Model}
We compare the proposed classification model with its variants for a justification of our design and a more comprehensive understanding of the model. 

\begin{table}[]
\caption{The performance of the variants of DCM classifier on the training set of ACDC}
\centering
\begin{tabular}{c c}
\hline
\noalign{\vskip 0.0in}
\multicolumn{1}{|c}{DCM Classifier Input} & \multicolumn{1}{|c|}{\# of Mis-Classification} \\ 
\multicolumn{1}{|c}{} & \multicolumn{1}{|c|}{on the 60 DCM, MINF} \\ 
\multicolumn{1}{|c}{} & \multicolumn{1}{|c|}{and NOR cases} \\ 
\hline
\multicolumn{1}{|c}{$\mathit{V_{LVC,ES}}$, $\mathit{RMD}$, $\mathit{TMD}$ (proposed)} & \multicolumn{1}{|c|}{\textbf{2}} \\ 
\hline
\multicolumn{1}{|c}{$\mathit{V_{LVC,ES}}$, $\mathit{TMD}$} & \multicolumn{1}{|c|}{\textbf{2}} \\ 
\hline
\multicolumn{1}{|c}{$\mathit{V_{LVC,ES}}$, $\mathit{RMD}$} & \multicolumn{1}{|c|}{3} \\ 
\hline
\multicolumn{1}{|c}{$\mathit{V_{LVC,ES}}$} & \multicolumn{1}{|c|}{4} \\ 
\hline
\end{tabular}
\end{table}

\subsubsection{Importance of Motion-Characteristic Features}
To better understand the value of the two proposed motion-characteristic features, we further train three variants of DCM classifier which use zero or one motion-characteristic feature as input. And the set of input features is the only difference between these models. As shown in Table 6, on the 60 cases of NOR, MINF and DCM, while DCM classifier makes only two errors, the variant using only shape-related feature $\mathit{V_{LVC,ES}}$ misclassifies 4 cases. But improvements can be made by using at least one motion-characteristic feature. As can be visualized in Fig.8, the motion characteristic features RMD and TMD allow the separation of the majority of the cases of DCM and MINF. Combining them with the shape-related feature $\mathit{V_{LVC,ES}}$ together as the input, the DCM classifier can make more accurate classification.

\begin{figure}[t]
\centering
\includegraphics[width=9.5cm, height=7.5cm]{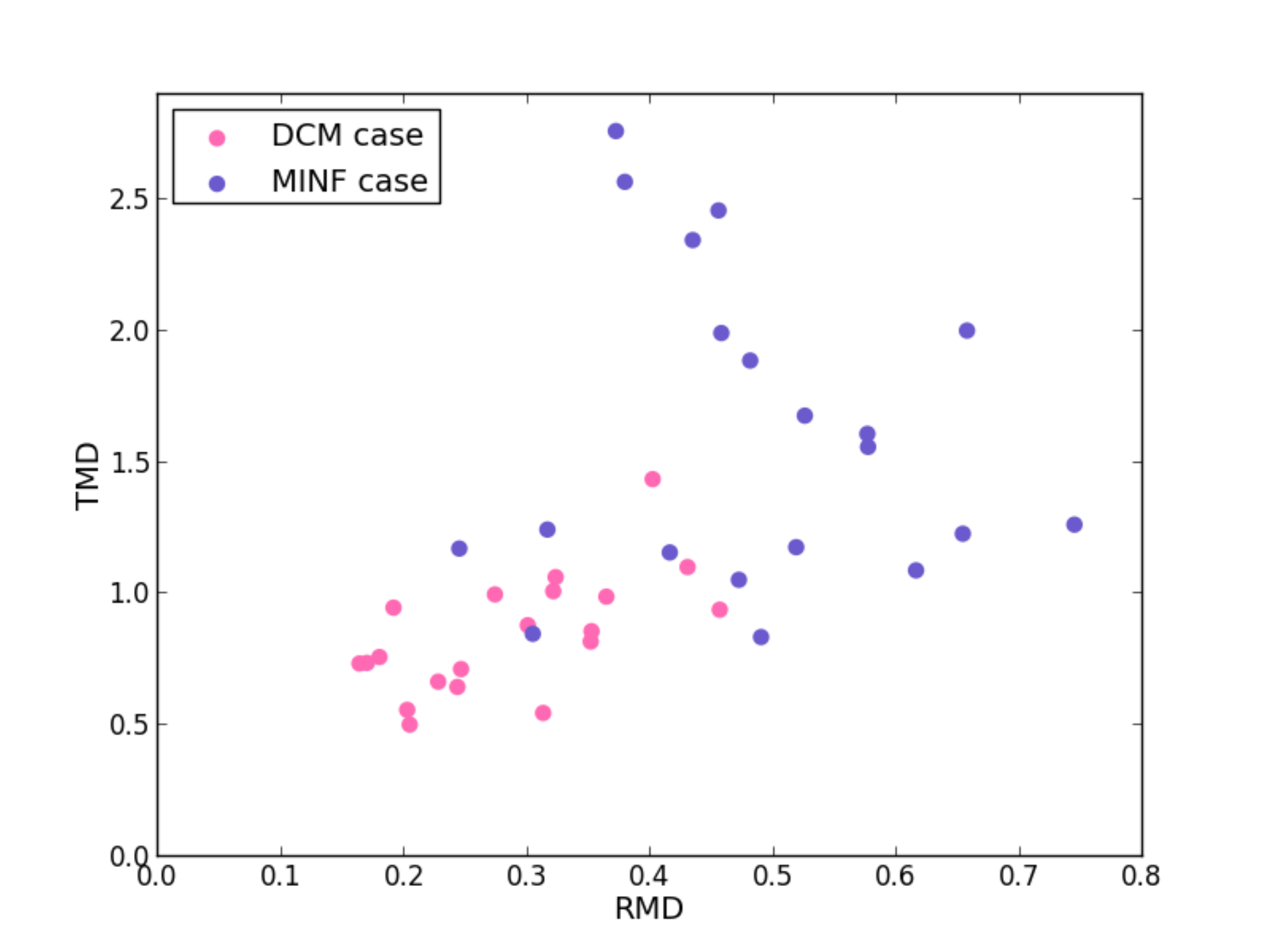}
\caption{The motion-characteristic features ($\bm{\mathit{RMD}}$ and $\bm{\mathit{TMD}}$) of the DCM and MINF cases in the training set of ACDC. The majority of the cases are well separable with these two features.}
\end{figure}

\begin{table}[t]
\caption{The 5-fold cross validation accuracy on the training set of ACDC of the variants of the proposed classification model}
\centering
\begin{tabular}{c cc}
\hline
\noalign{\vskip 0.0in}
\multicolumn{1}{|c}{Method} & \multicolumn{1}{|c}{BSA- } & \multicolumn{1}{|c|}{Non-} \\ 
\multicolumn{1}{|c}{} & \multicolumn{1}{|c}{Normalized } & \multicolumn{1}{|c|}{Normalized} \\ 
\multicolumn{1}{|c}{} & \multicolumn{1}{|c}{Features} & \multicolumn{1}{|c|}{Features} \\ 
\hline
\multicolumn{1}{|c}{logistic regression classifiers} & \multicolumn{1}{|c}{\textbf{95\%}} & \multicolumn{1}{|c|}{\textbf{94\%}}  \\ 
\multicolumn{1}{|c}{(proposed model)} & \multicolumn{1}{|c}{} & \multicolumn{1}{|c|}{}  \\ 
\hline
\multicolumn{1}{|c}{logistic regression classifiers} & \multicolumn{1}{|c}{63\%} & \multicolumn{1}{|c|}{64\%}  \\ 
\multicolumn{1}{|c}{in inversed order} & \multicolumn{1}{|c}{} & \multicolumn{1}{|c|}{}  \\ 
\hline
\multicolumn{1}{|c}{Lasso classifiers} & \multicolumn{1}{|c}{89\%} & \multicolumn{1}{|c|}{91\%}  \\ 
\hline
\multicolumn{1}{|c}{LassoCV classifiers} & \multicolumn{1}{|c}{80\%} & \multicolumn{1}{|c|}{81\%}  \\ 
\hline
\multicolumn{1}{|c}{random forest classifiers} & \multicolumn{1}{|c}{85\%} & \multicolumn{1}{|c|}{87\%}  \\ 
\hline
\multicolumn{1}{|c}{logistic regression classifiers} & \multicolumn{1}{|c}{88\%} & \multicolumn{1}{|c|}{88\%}  \\ 
\multicolumn{1}{|c}{w/o manual feature selection} & \multicolumn{1}{|c}{} & \multicolumn{1}{|c|}{}\\ 
\hline
\multicolumn{1}{|c}{SVM classifiers} & \multicolumn{1}{|c}{87\%} & \multicolumn{1}{|c|}{84\%}  \\ 
\multicolumn{1}{|c}{w/o manual feature selection} & \multicolumn{1}{|c}{} & \multicolumn{1}{|c|}{}  \\ 
\hline
\multicolumn{1}{|c}{RVM classifiers} & \multicolumn{1}{|c}{88\%} & \multicolumn{1}{|c|}{72\%}  \\ 
\multicolumn{1}{|c}{w/o manual feature selection} & \multicolumn{1}{|c}{} & \multicolumn{1}{|c|}{}  \\ 
\hline
\multicolumn{1}{|c}{Lasso classifiers} & \multicolumn{1}{|c}{85\%} & \multicolumn{1}{|c|}{86\%}  \\ 
\multicolumn{1}{|c}{w/o manual feature selection} & \multicolumn{1}{|c}{} & \multicolumn{1}{|c|}{}  \\ 
\hline
\multicolumn{1}{|c}{LassoCV classifiers} & \multicolumn{1}{|c}{84\%} & \multicolumn{1}{|c|}{87\%}  \\ 
\multicolumn{1}{|c}{w/o manual feature selection} & \multicolumn{1}{|c}{} & \multicolumn{1}{|c|}{}  \\ 
\hline
\multicolumn{1}{|c}{random forest classifiers} & \multicolumn{1}{|c}{86\%} & \multicolumn{1}{|c|}{88\%}  \\ 
\multicolumn{1}{|c}{w/o manual feature selection} & \multicolumn{1}{|c}{} & \multicolumn{1}{|c|}{}  \\ 
\hline
\multicolumn{1}{|c}{HDDA classifiers} & \multicolumn{1}{|c}{49\%} & \multicolumn{1}{|c|}{46\%}  \\ 
\multicolumn{1}{|c}{w/o manual feature selection} & \multicolumn{1}{|c}{} & \multicolumn{1}{|c|}{}  \\ 
\hline
\multicolumn{1}{|c}{one single random forest} & \multicolumn{1}{|c}{87\%} & \multicolumn{1}{|c|}{88\%}  \\ 
\multicolumn{1}{|c}{w/o binary classification} & \multicolumn{1}{|c}{} & \multicolumn{1}{|c|}{}  \\ 
\hline
\multicolumn{1}{|c}{one single MLP} & \multicolumn{1}{|c}{84\%} & \multicolumn{1}{|c|}{84\%}  \\ 
\multicolumn{1}{|c}{w/o binary classification} & \multicolumn{1}{|c}{} & \multicolumn{1}{|c|}{}  \\ 
\hline

\end{tabular}
\end{table}

\subsubsection{Proposed Model on Non-Normalized Features}
We test whether BSA normalization is required for our model to achieve high performance. Among the 9 proposed features, only the values of $\mathit{V_{RVC,ED}}$ and $\mathit{V_{LVC,ES}}$ would be different without BSA-normalization. And only RVA and MINF classifiers which use these two features as input would be affected. As presented in Table 7, without BSA-normalization on the features, the 5-fold cross validation accuracy on ACDC training set only drops a little bit to 94\%. The proposed model still remains accurate.

\subsubsection{Proposed Model with Inversed Classifier Order}
As presented previously, the 4 classifiers in the proposed model are arranged according to the estimated difficulties of the corresponding classification task. To confirm the importance of this order, we create another model by inversing the order of the classifiers. So, unlike the proposed model shown in Fig.2, in this variant, a case goes through successively MINF, DCM, HCM and RVA classifiers instead. As shown in Table 7, the accuracy of this variant is quite low. Hence the proposed order of the classifiers is indeed important.

\subsubsection{Variants with Other Classifier Models}
We replace the proposed logistic regression classifiers with 3 other types of classifiers on the same sets of input features, including Lasso, LassoCV (Lasso with model selection by cross-validation) and random forest. Details of these models are available in Appendix C. As reported in Table 7, their performances are clearly below that of the proposed model. Our choice of logistic regression as the classifier model is hence justified.

\subsubsection{Variants without Manual Input Feature Selection}
To evaluate if the manual feature selection is useful for the model to be accurate, we train several modified versions of the proposed model without manual feature selection. They all consist of 4 classifiers to perform the same binary classification tasks as in the proposed model. But each of the 4 classifiers of these variants takes all 9 features together as input. In total, we implement 6 models with the following models as their classifiers respectively (details of these models are available in Appendix C): support vector machine (SVM), relevance vector machine (RVM), Lasso, LassoCV, random forest and high dimensional discriminant analysis (HDDA) model.

As reported in Table 7, on the BSA-normalized features as well as on the non-normalized features, they all have accuracy lower than that of the proposed model by at least 6\%. This justifies the necessity of manual feature selection, at least on a relatively small dataset like ACDC. We are not yet able to examine the importance of manual feature selection on large datasets.

To better understand the roles of the features, we further examine the variant with random forest classifiers without manual feature selection trained on the 100 cases of ACDC training set. For each of the 4 classifiers, we compute the feature importance for each of the 9 features to determine the most important one. The importance of a feature is defined as the total reduction of the entropy brought by that feature in all the trees in the random forest. We find that for RVA, HCM and DCM classifiers, the most important features are $\mathit{R_{RVCLV,ED}}$, $\mathit{R_{LVMLVC,ED}}$ and $\mathit{V_{LVC,ES}}$ respectively, which are among the features manually selected for the corresponding proposed classifiers. For MINF classifier, the two most important features are $\mathit{R_{RVCLV,ED}}$ and $\mathit{EF_{LVC}}$, which have roughly the same importance (0.35 and 0.32). Only $\mathit{EF_{LVC}}$ is used in the proposed model according to its direct relevance. These observations provide further support for our manual feature selection.

\subsubsection{Variants without Binary Classification}
The proposed model divides the 5-category classification task into 4 binary classification sub-tasks. In order to understand whether this special design contributes to the achieved high accuracy, we train and evaluate 2 variants on the same set of 9 features. A random forest and a multi-layer perceptron (MLP) are respectively trained to predict a case to be one of the five categories directly without binary classification (details of these models are available in Appendix C). As reported in Table 7, their performances are not as good as that of the proposed model. Hence the strategy of performing a series of binary classification makes sense.

\section{Conclusion and Discussion}

We propose a method of cardiac pathology classification based on originally designed and trained neural networks and classifiers \footnote{The code and the model will be available in this repository: https://github.com/julien-zheng/CardiacMotionFlow}. A novel semi-supervised training method is applied to train ApparentFlow-net which provides pixel-wise motion information. Combining the apparent flow generated by ApparentFlow-net and the segmentation masks predicted by LVRV-net, we introduce two novel features that characterize the motion of myocardial segments. These motion-characteristic features are not only intuitive for visualization but also very valuable in classification. The proposed classification model consists of 4 small binary classifiers. Each classifier works independently and takes up to 3 features with clearly explainable relevance as input. On ACDC training set and testing set, the proposed model achieves 95\% and 94\% respectively as classification accuracy. Its performance is hence comparable to that of the state-of-the-art. To justify our design of the proposed classification model, we also quantitatively compare it with other models.

The apparent flow generated by ApparentFlow-net and the originally designed time series of myocardial segment motion are not only straightforward to understand but also useful for classification. We believe that making the automatic methods more understandable and explainable is important, as it is not only helpful to facilitate the implementation and application of the research of medical image analysis in clinics but also useful to improve transparency and to gain trust in medical practice (\cite{Holzinger:2017}, \cite{Rueckert:2016}).

Furthermore, the motion information we extract from the apparent flow is fairly rich. We believe that ApparentFlow-net may be a powerful tool of motion extraction for the community. The way we extract the time series and the motion-characteristic features from the flow maps is just one of the so many possible applications. Also, ApparentFlow-net is trained in a semi-supervised manner. This training approach is highly relevant to the current situation of data availability in medical image analysis, as we usually have access to a relatively large amount of unlabeled data and a relatively small amount of labeled data. In a word, much more potential applications in various circumstances of apparent flow are yet to be explored.

Regarding the extraction of the motion-characteristic features, one could use the segmentation network to segment all frames and then derive the motion-characteristic features from the segmentation masks. However, we find that the resulting time series characterizing the cardiac motion (e.g. the time series of the radius and thickness as shown in the second column of Fig.6) by this approach are not as temporally consistent as we would expect. In fact, the segmentation network was trained to segment the frames at ED and ES only. And no constraint has ever been imposed to make the segmentation masks temporally consistent. The problem would be clearer if we look at the two frames in the first column in Fig.5. While the ED frame (upper image) is easy to segment, the other frame (lower image) appears to be more challenging due to the presence of massive trabeculations. Moreover, as the segmentation network segments the two frames independently, it is not obvious how to ensure the consistency of the segmentation masks. This problem can be solved using the ApparentFlow-net instead to extract motion. As shown in the second column of Fig.6, with the ApparentFlow-net, the extracted time series of the radius and thickness of the segments are reasonably smooth, which reflects the enforced temporal consistency.

We could have used existing traditional registration models to supervise the training of ApparentFlow-net or even replace ApparentFlow-net by a deformable registration algorithm (e.g. LDDMM, LCC-Demons, etc.). However, we notice that in order to make the traditional registration models work reasonably well on an unseen dataset like ACDC, the estimation and finetuning of key parameters in these models are necessary. For instance, the authors of \cite{Krebs:2019} empirically estimate the key parameters of LCC-Demons (\cite{Lorenzi:2013}) and SyN (\cite{Avants:2008}) before applying them on ACDC. Our method is simpler in the sense that it learns everything from data and requires no manual model/parameter estimation/adjustment. Hence, on the one hand, our method is easier and more convenient to be applied to various datasets that are reasonably similar to the training set. On the other hand, it allows us to take advantage of the increasing number of data available in the community. We believe a method with these advantages is very interesting and worth trying. Moreover, as far as we observe, our registration method is accurate enough to characterize the cardiac motion. Some examples are provided in Fig.D.9 to show that the generated apparent flow characterizes the motion patterns of typical cases in several pathological categories. And as shown in the second column of Fig.6, with the apparent flow generated by the ApparentFlow-net, the extracted time series of the radius and thickness of the segments enable us to easily distinguish the typical cases of different cardiac pathologies.

A straightforward comparison with prior works on 2D registration methods on the ACDC dataset shows that the ApparentFlow-net performs rather well. Indeed when  looking at the Dice coefficients achieved on LVC and RVC, our approach leads to 0.94 and 0.87 respectively (see Table 3). In \cite{Hering:2019}, the authors describe a learning-based method leading to Dice coefficients at best equal to 0.90 on the same structures (based on Fig.3 of \cite{Hering:2019}) and also performances of a non-learning-based method similar to \cite{Ruhaak:2013} with at most 0.80 of Dice. Note however that in this comparison, differences on cross-validation (5-fold v.s. 10-fold), slice selection and ROI determination may hinder the analysis.

While analyzing and extracting the cardiac motion, we adopt a 2D slice-by-slice processing method, without taking the motion on neighboring slices into account. The reason behind this choice is the fact that the inter-slice distance in the short-axis MRIs in ACDC is quite large. Usually, the inter-slice distance between two adjacent slices in MRI stacks is 5 to 10mm. The heart may hence have obviously different shape and motion even on two adjacent slices. In this case, ignoring the neighboring slices for motion estimation might be reasonable. However, if our method is to be applied on some volumetric images with small inter-slice distance, a modification of the approach by taking neighboring slices into account might be beneficial.

An issue that would hinder the generalization of pathology classification models like ours is the lack of a standard and universal definition of pathological category \cite{Suinesiaputra:2016}. For instance, there is another public dataset made available for the MICCAI 2009 challenge on automated LV segmentation \cite{Radau:2009} (the dataset is also known as the Sunnybrook dataset (SD)) containing pathological cases. The 4 categories of SD are heart failure with infarction, heart failure without infarction, LV hypertrophy and healthy. While a hypertrophic case in ACDC has a LV cardiac mass over $110\mathit{g/m^2}$ and several myocardial segments of thickness over 15mm at ED by definition, the hypertrophic cases according to SD definition only need to have a LV cardiac mass over $83\mathit{g/m^2}$. And no threshold is proposed for the myocardial segment thickness by the SD definition. In fact, we find multiple cases in SD which are of LV cardiac mass between $83\mathit{g/m^2}$ and $110\mathit{g/m^2}$ and maximal segment thicknesses well below 15mm. They are identified as hypertrophic cases in SD. But they would not be considered as hypertrophic at all according to ACDC. Similarly, the category of infarction is defined differently in SD and ACDC. In SD, the infarction is determined by the evidence of late gadolinium enhancement; abnormal cardiac motion might not be observable. Yet in ACDC, the infarction category is defined by the presence of abnormal motion. With such discrepancies between the definitions in different datasets, it is difficult for the community to train a classification model on a dataset such that it generalizes well to the others. We hence appeal for more attention on this issue.

Another issue that may limit the generalization of our classification model is the small size of the ACDC dataset used for training. ACDC training set has only 100 cases of 5 pathological categories. Moreover, in each pathological category, there are only 20 cases. Consequently, on the one hand, many pathological categories are not included in ACDC. On the other hand, for each of the 5 pathological categories in ACDC, we would expect that the 20 cases might not be enough to represent all cases in the category. In order to achieve good generalization, we may need larger datasets with more pathological categories to train the model.

Also, notice that the proposed simple classification model of only 14 parameters is somewhat specific to the ACDC dataset. If we need to adapt our model to perform classification on a larger dataset with more pathological categories, it may be necessary to increase the size and hence the number of parameters of the model.

Finally, we would like to point out that although some single-value hand-crafted motion-characteristic features (e.g. $\mathit{RMD}$ and $\mathit{TMD}$) are used in this paper, we believe that for some pathology it would be better to use the whole time series of segment radius or thickness as input to a classification model. For instance, if we aim to discover subtler characteristics related to the motion (e.g. dyssynchrony, septal flash) from a larger dataset, doing so might become appropriate and necessary. We expect to carry out research on this topic in the future.

\section*{Acknowledgments}

%{\footnotesize The authors acknowledge the partial support from the European Research Council (MedYMA ERC-AdG-2011-291080).}
The authors acknowledge the partial support from the European Research Council (MedYMA ERC-AdG-2011-291080).

%%%%%%%%%%%%%%%%%%%%%%%%%%%%%%%%%%%%%%%%%%
%\section*{References}

%%Harvard
\bibliographystyle{model2-names}\biboptions{authoryear}
\bibliography{refs}

%%%%%%%%%%%%%%%%%%%%%%%%%%%%%%%%%%%%%%%%%%%%
\appendix

\section{Loss Function for Training ApparentFlow-Net}

To penalize the crossing or large rotations of flows, we compute the difference between of the warped x-components (resp. y-components) of each pair of horizontally (resp. vertically) adjacent pixels $\bm{P^{x+}}$ and $\bm{P}$ (resp. $\bm{P^{y+}}$ and $\bm{P}$). There is a crossing if and only if this difference is smaller than 0, for which a penalty which is equal to the square of this difference applies. Otherwise no penalty applies. Hence we come up with the term $L_{\mathit{CROSS}}(\bm{F_t})$ to penalize the crossing of flows:
\begin{equation}
\begin{split}
& \:L_{\mathit{CROSS}}(\bm{F_t}) \\
= &\sum_{\bm{P}} \mathit{min}\Big(\big(x+1+F_{t}^x(\bm{P^{x+}})\big)-\big(x+F_{t}^x(\bm{P})\big), 0\Big)^2 \\
&+ \sum_{\bm{P}} \mathit{min}\Big(\big(y+1+F_{t}^y(\bm{P^{y+}})\big)-\big(y+F_{t}^y(\bm{P})\big), 0\Big)^2 \\
= &\sum_{\bm{P}} \mathit{min}(1+\frac{\partial F_{t}^x(\bm{P})}{\partial x}, 0)^2 + \mathit{min}(1+\frac{\partial F_{t}^y(\bm{P})}{\partial y}, 0)^2
\end{split}
\end{equation} 
in which $\partial F_{t}^x(\bm{P}) / \partial x$ is computed with finite difference as $F_{t}^x(\bm{P^{x+}}) - F_{t}^x(\bm{P})$ (and similarly for $\partial F_{t}^y(\bm{P}) / \partial y$).

The Dice function in the term $L_{GT}(\bm{F_{ES}})$ is defined on two images $U$ and $V$ as described in \cite{Zheng:2018}
\begin{equation}
\mathit{Dice}(U,V) = -\frac{ 2\sum_{\bm{P}} U(\bm{P})V(\bm{P}) + \epsilon}{ \sum_{\bm{P}} U(\bm{P}) + \sum_{\bm{P}} V(\bm{P}) + \epsilon}
\end{equation}
with $\epsilon=1$ a term for better numerical stability in training.

\section{Variants of the Proposed Classification Model with Different Values of Parameter $C$}
We also perform 5-fold cross-validation on the ACDC training set for the variants of the proposed classification model by varying the parameter $C$ in the 4 logistic regression classifiers. Their performances are reported in Table B.8.

\begin{table}[t]
\caption{The 5-fold cross-validation performance on the ACDC training set of some variants of the proposed classification model with various values of parameter $C$}
\centering
\begin{tabular}{c c}
\hline
\noalign{\vskip 0.0in}
\multicolumn{1}{|c}{$C$} & \multicolumn{1}{|c|}{Training Set Accuracy} \\ 
\hline
\multicolumn{1}{|c}{1} & \multicolumn{1}{|c|}{76\%} \\ 
\hline
\multicolumn{1}{|c}{5} & \multicolumn{1}{|c|}{88\%} \\ 
\hline
\multicolumn{1}{|c}{10} & \multicolumn{1}{|c|}{92\%} \\ 
\hline
\multicolumn{1}{|c}{50} & \multicolumn{1}{|c|}{95\%} \\ 
\hline
\multicolumn{1}{|c}{100} & \multicolumn{1}{|c|}{95\%} \\ 
\hline
\multicolumn{1}{|c}{500} & \multicolumn{1}{|c|}{93\%} \\ 
\hline
\multicolumn{1}{|c}{1000} & \multicolumn{1}{|c|}{93\%} \\ 
\hline
\multicolumn{1}{|c}{5000} & \multicolumn{1}{|c|}{93\%} \\ 
\hline
\end{tabular}
\end{table}

\section{Variants of the Proposed Classification Model with Different Classifiers and Input Features}

\subsection{Variants with Other Classifier Models}
We replace the proposed ridge logistic regression classifiers by other types of classifiers on the same sets of input features:\\
\textbullet \ Lasso classifiers: in this variant, each of the 4 classifiers is a least absolute shrinkage and selection operator (Lasso). The constant $\mathit{alpha}$ that multiplies the $L_1$ term is empirically chosen to be $10^{-4}$.\\
\textbullet \ LassoCV classifiers: each of the 4 classifiers is a Lasso model with model selection by cross-validation (LassoCV). The optimal constant $\mathit{alpha}$ is searched in the range $[10^{-4}, 10^{-0.5}]$ in a 4-fold cross-validation on the training data.\\
\textbullet \ random forest classifiers: each of the 4 classifiers is a random forest of 1000 trees which expand their nodes in training until all leaves are pure or all leaves contain less than 2 samples. Entropy is used to measure the quality of a split in training.

\subsection{Variants without Manual Input Feature Selection}
We train several variants of the proposed model without manual feature selection. They all consist of 4 classifiers arranged in the same order to perform the same binary classification tasks as in the proposed model. But each of the 4 classifiers in these variants takes all the 9 features together as input. In total, we implement and examine 6 variants with the following models as their classifiers respectively:\\
\textbullet \ Variant with SVM classifiers: each of the 4 binary classifiers is a support vector machine (SVM) with linear kernel and penalty parameter C=50. \\
\textbullet \ Variant with RVM classifiers: each of the 4 binary classifiers is a relevance vector machine (RVM) as introduced in \cite{Tipping:2003} with linear kernel. \\
\textbullet \ Variant with Lasso classifiers: each of the 4 binary classifiers is a Lasso. Lasso is known as a model capable of performing both variable selection and regularization. $\mathit{alpha}$, the constant that multiplies the $L_1$ term, is empirically set to $10^{-4}$. \\
\textbullet \ Variant with LassoCV classifiers: each of the 4 binary classifiers is a Lasso with model selection in a 4-fold cross-validation on the training data. The optimal constant $\mathit{alpha}$ is searched in the range $[10^{-4}, 10^{-0.5}]$.\\
\textbullet \ Variant with random forest classifiers: in this variant, each of the 4 binary classifiers is a random forest of 1000 trees which expand their nodes in training until all leaves are pure or all leaves contain less than 2 samples. Entropy is used to measure the quality of a split in training.\\
\textbullet \ Variant with HDDA classifiers: each of the 4 binary classifiers is a high dimensional discriminant analysis (HDDA) model, which is an expectation-maximization algorithm designed for high-dimensional data clustering based on the ideas of dimension reduction and parsimonious modeling (\cite{Bouveyron:2007}, \cite{Orlhac:2018}). Though the 9-feature space in this paper is not high dimensional, we show the performance of such a sophisticated method for comparison.\\

\subsection{Variants without Binary Classification}
We train and evaluate the following 2 variants on all the 9 input features. These variants are obtained by replacing the 4 binary classifiers with a single multi-class one:\\
\textbullet \ Variant using random forest: it is a random forest of 1000 trees which expand their nodes in training until all leaves are pure or all leaves contain less than 2 samples. Entropy is used to measure the quality of a split in training.\\
\textbullet \ Variant using MLP: it is a multi-layer perceptron (MLP). It has 2 hidden layers of 32 neurons with tanh activation function. Adam optimizer is used to train it for $10^5$ epochs with learning rate 0.001.\\

\subsection{Implementation of the Variants}
Among the above variants of the proposed classification model with different classifiers and input features, the HDDA classifiers are implemented using the HDDA python toolbox downloaded from the GitHub page \url{https://github.com/mfauvel/HDDA}, the RVM classifiers are implemented in Python according to the method described in \cite{Tipping:2003}, and all the other variants are implemented with Scikit-learn.

\section{Examples of Apparent Flow Generated by the ApparentFlow-net}
\begin{figure*}[t]
\centering
\includegraphics[width=16.0cm, height=18.8cm]{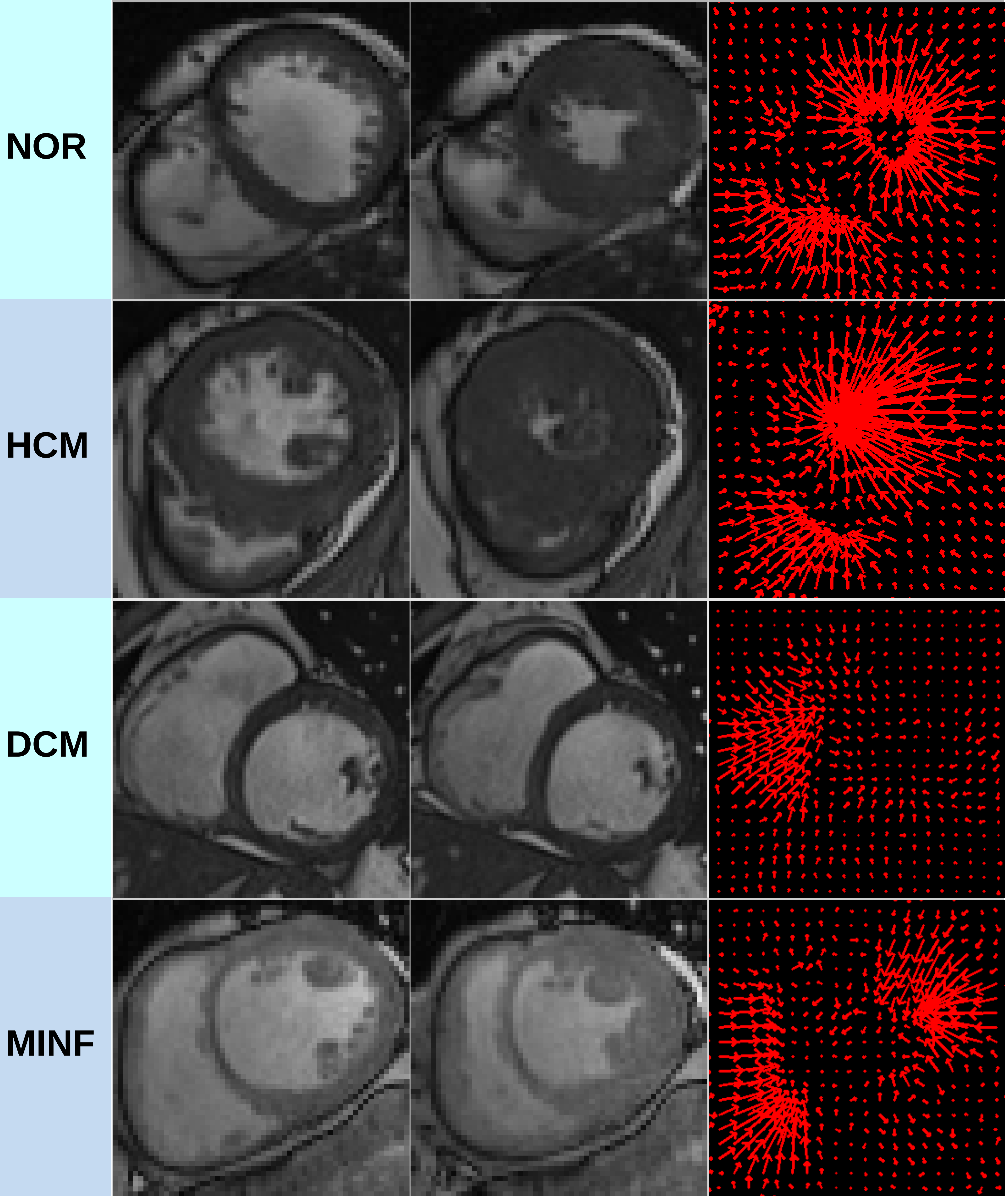}
\caption{Four examples of the apparent flow generated by the ApparentFlow-net of 4 ACDC training set cases in different pathological categories. The images in the first column are the frames at ED. The images in the second column are the frames around ES. The apparent flow maps corresponding to the pairs of frames in the first and second columns are shown in the third column. The apparent flow can indeed characterize the cardiac motion of the typical cases in the pathological categories. NOR: synchronous and uniform flow on LVM; HCM: excessively large flow on LVM; DCM: very small flow on LVM; MINF: asynchronous and ununiform flow on LVM.}
\end{figure*}

We provide 4 examples of the apparent flow generated by the ApparentFlow-net of 4 ACDC training set cases in different pathological categories. In Fig.D.9, given the frames at ED (first column) and the frames around ES (second column), we apply the trained ApparentFlow-net to generate the apparent flow maps (third column).

Visually, we find that the apparent flow can indeed characterize the cardiac motion of the typical cases in the pathological categories. As expected, the apparent flow on the LVM of a NOR case is oriented along the gradient of the image intensity and has roughly the same amplitude throughout the left ventricle, signifying the synchronous and uniform contraction and thickening of the LVM of the NOR case. For a HCM cases, we can see that the flow on LVM is excessively large, which means that the contraction and thickening is excessive, a typical phenomenon we find on HCM cases. Conversely, the flow on the LVM of a DCM case is small since the hearts of DCM cases usually do not contract or thicken enough. Finally, the flow on the LVM of a MINF case is not uniform: some myocardial segments contract and thicken much less than the others. This is a typical symptom that we can find on MINF cases.

\end{document}